\documentclass[11pt,a4paper]{article}
\usepackage{times,latexsym}
\usepackage{url}
\usepackage[T1]{fontenc}

\usepackage[acceptedWithA]{tacl2021v1}

\usepackage{xspace,mfirstuc,tabulary}

\newif\iftaclinstructions
\taclinstructionsfalse 
\iftaclinstructions
\renewcommand{\confidential}{}
\renewcommand{\anonsubtext}{(No author info supplied here, for consistency with
TACL-submission anonymization requirements)}
\newcommand{\instr}
\fi

\iftaclpubformat 

\else

\fi

\usepackage{hyperref}
\usepackage{url}
\usepackage{graphicx}

\usepackage{booktabs}
\usepackage{subcaption}
\usepackage{multirow}
\usepackage{amssymb}
\usepackage{pifont}

\usepackage{amsmath} 
\usepackage{amsthm}
\usepackage{amssymb}
\usepackage{bm}
\usepackage{dsfont}
\usepackage{enumitem}
\usepackage{tikz}
\usetikzlibrary{matrix,positioning,decorations.pathreplacing}

\DeclareMathOperator*{\argmin}{arg\,min}
\DeclareMathOperator{\cov}{\mathbb{C}ov}
\DeclareMathOperator{\E}{\mathbb{E}}
\DeclareMathOperator{\Iden}{\mathbb{I}}
\DeclareMathOperator{\Real}{\mathbb{R}}
\newcommand{\mat}[1]{\bm{#1}}

\makeatletter
\newtheorem*{rep@theorem}{\rep@title}
\newcommand{\newreptheorem}[2]{%
\newenvironment{rep#1}[1]{%
 \def\rep@title{#2 \ref{##1}}%
 \begin{rep@theorem}}%
 {\end{rep@theorem}}}
\makeatother

\newtheorem{theorem}{Theorem}
\newreptheorem{theorem}{Theorem}

\newtheorem{definition}{Definition}

\usepackage{adjustbox}

\definecolor{carminepink}{rgb}{0.92, 0.3, 0.26}
\definecolor{jade}{rgb}{0.0, 0.66, 0.42}

\definecolor{schema_red}{HTML}{B85450}
\definecolor{schema_green}{HTML}{82B366}
\definecolor{schema_celadon}{HTML}{0E8088}

\definecolor{schema_blue}{HTML}{2787BD}
\definecolor{schema_orange}{HTML}{D2A178}

\newcommand{\fact}[0]{\emph{factual }}
\newcommand{\stereo}[0]{\emph{stereotypical }}
\newcommand{\empiric}[0]{\emph{empirical }}

\newcommand{\dama}[0]{\emph{DAMA}}
\newcommand{\ddama}[0]{\emph{2DAMA}}
\newcommand{\llama}[0]{Llama}
\newcommand{\leace}[0]{\emph{LEACE}}
\newcommand{\dd}[0]{\emph{Dual Debiasing}}

\definecolor{bottlegreen}{rgb}{0.0,0.42,0.31}
\definecolor{babypink}{rgb}{0.83,0.42,0.54}
\definecolor{royalpurple}{rgb}{.471,.318,.663}

\title{
Dual Debiasing: Remove Stereotypes and Keep Factual Gender for \\ Fair Language Modeling and Translation}

\author{Tomasz Limisiewicz \and David Mare\v{c}ek \and Tom\'{a}\v{s} Musil \\
    Institute of Formal and Applied Linguistics, Faculty of Mathematics and Physics \\
    Charles University, Prague, Czech Republic \\
  \texttt{\{limisiewicz, marecek, musil\}@ufal.mff.cuni.cz}
}

\date{}

\begin{document}
\maketitle

\begin{abstract}

Mitigation of biases, such as language models' reliance on gender stereotypes, is a crucial endeavor required for the creation of reliable and useful language technology.
The crucial aspect of debiasing is to ensure that the models preserve their versatile capabilities, including their ability to solve language tasks and equitably represent various genders. 
To address this issue, we introduce streamlined \textit{Dual Dabiasing Algorithm through Model Adaptation} (\ddama{}).
Novel \dd{} enables robust reduction of stereotypical bias while preserving desired factual gender information encoded by language models.
We show that \ddama{} effectively reduces gender bias in English and is one of the first approaches facilitating the mitigation of stereotypical tendencies in translation.
The proposed method's key advantage is the preservation of factual gender cues, which are useful in a wide range of natural language processing tasks.

\end{abstract}

\section{Introduction}

Gender representation in large language models (LLMs) has been the topic of significant research effort \cite{stanczak_survey_2021, kotek-etal-2023-gender}.
Past studies have predominantly focused on such representation to identify and mitigate social biases.
Admittedly, biases are a challenging issue limiting the reliability of LLMs in real-world applications. 
Yet, we argue that preserving particular types of gender representation is crucial for fairness and knowledge acquisition in language models.

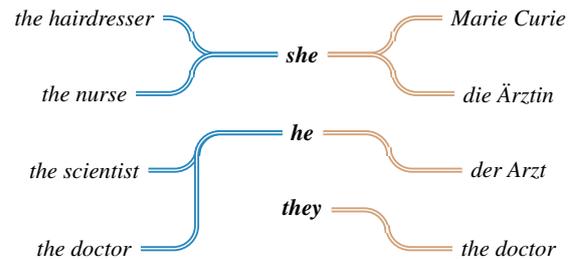
\begin{figure}[t]
    \centering
\begin{adjustbox}{max width=\linewidth}
\begin{tikzpicture}
    \tikzset{
        every node/.style={font=\small},
        column 1/.style={anchor=center},
        column 2/.style={anchor=center},
        column 3/.style={anchor=center}
    }

    \matrix (m) [
        matrix of nodes,
        nodes in empty cells,
        column sep=1.7cm,
        row sep=0.06cm
    ] {
        \textit{the hairdresser} &  & \textit{Marie Curie} \\
        & \textit{\textbf{she}} & \\
        \textit{the nurse} &  & \textit{die Ärztin} \\
        & \textit{\textbf{he}}& \\
        \textit{the scientist} &  & \textit{der Arzt} \\
        & \textit{\textbf{they}} & \\
        \textit{the doctor} &  & \textit{the doctor}\\
    };

    \draw[double, thick, rounded corners=10pt, draw=schema_blue] (m-1-1.east) -- ++(0.435,0) |- (m-2-2.west);
    \draw[double, thick, rounded corners=10pt, draw=schema_blue] (m-3-1.east) -- ++(0.85,0) |-  (m-2-2.west);
    \draw[double, thick, rounded corners=10pt, draw=schema_orange] (m-2-2.east) -- ++(1.0,0) |- (m-1-3.west);
    \draw[double, thick, rounded corners=10pt, draw=schema_orange] (m-2-2.east)  -- ++(1.0,0) |-  (m-3-3.west);
    
    \draw[double, thick, rounded corners=10pt, draw=schema_blue] (m-5-1.east) -- ++(0.75,0) |- (m-4-2.west);
    \draw[double, thick,rounded corners=10pt, draw=schema_blue] (m-7-1.east) -- ++(0.85,0) |- (m-4-2.west);
    \draw[double, thick, rounded corners=10pt,draw=schema_orange] (m-4-2.east) -- ++(1.0,0) |-  (m-5-3.west);

    \draw[double, thick, rounded corners=10pt, draw=schema_orange] (m-6-2.east) -- ++(0.85,0) |-  (m-7-3.west);

\end{tikzpicture}
\end{adjustbox}
    \caption{Dual character of gender signals encoded in language models:
    \textcolor{schema_blue}{stereotypical} cues are shown on the left, and \textcolor{schema_orange}{factual} gender cues are shown on the right-hand side.
    ``\textit{Die Ärztin}'' and ``\textit{der Arzt}'' are respectively female and male German translation for ``\textit{the doctor}''.}
    \label{fig:figure-one}
\end{figure}

To provide a more detailed perspective, we draw examples of both unwanted and beneficial types of gender signals in LLMs.
Undesirable biases are typically inherited from stereotypes and imbalances in the training corpora and tend to be further amplified during the model training \citep{van-der-wal-etal-2022-birth, gallegos-etal-2024-bias}.
Biases are manifested in multiple ways, including unequal representation (models are more likely to generate mentions of a specific overrepresented gender), stereotypical associations (particular contexts are associated with one gender based on stereotypical cues, e.g., ``\textit{politics and business are male domains}'', while ``\textit{family is a female domain}'').
It has been shown that, due to bias, LLMs struggle with high-stakes decision-making and are prone to produce discriminatory predictions.
Examples of such a sensitive application are the automatic evaluation of CVs and biographical notes \citep{de-arteaga_bias_2019}, where some professions are stereotypically associated with a specific gender.
Therefore, individuals of that gender could benefit from unfair advantage when assessed by an LLM-based evaluator.

Nevertheless, LLMs should understand and represent gender signals.
For instance, chatbots should be persistent in addressing the user with their preferred gender pronouns after they are revealed \cite{limisiewicz-marecek-2022-dont}.
Adequate representation of gender is also required for knowledge acquisition, for example, in question answering (QA), to correctly answer ``\textit{Maria Skłodowska-Curie}'' to the question ``\textit{Who was the first woman to win a Nobel Prize?}''.
Gender sensitivity is even more critical in morphologically rich languages, where gender mentions are much more ubiquitous, e.g., through morphological markings (as in German, Czech, or Russian) \cite{hellinger2002gender}.
Examples of dual characters of gender encoding are shown in Figure~\ref{fig:figure-one}.

\subsection{Solution}

To address these intricate ways gender signals are present in natural language, we introduce a new method \ddama{} that post-hoc modifies pre-trained language models to represent gender in an equitable way, i.e., without stereotypical bias but with factual gender information. 
The approach leverages recent effective methods for model adaptation \citet{limisiewicz2024debiasing, meng_locating_2023}, and concept erasure \citet{belrose2023leace}.
\textbf{As the core contribution, we introduce the novel method of \dd{} that aims at our core problem of decreasing bias while keeping equitable factual gender representation}.

We present a strong theoretical backing of the effectiveness of our new method \ddama{} in Section~\ref{sec:methodology}.
Subsequently, we show the experimental approach in the results for applying the method to the aforementioned case of gender: stereotypical vs. factual (Sections~\ref{sec:exp-setting},~\ref{sec:results-lm}).
Furthermore, we are the first to mitigate LLMs' multilingual gender bias manifested in translation from English to three morphologically rich languages, presented in Section~\ref{sec:results-mt}.

\subsection{Research Questions}

In the experimental part, we present an extensive analysis of the observed patterns in gender representation in four models and the effect of different design choices in the debiasing approach.
The analysis specifically answers the research questions, grouped into four areas: 
\begin{enumerate}[itemsep=2mm]
    \item[A.] How extensive the update of parameters should be to erase gender bias? Can the extent of the update be implicitly learned by the debiasing method?
    \item[B.] To what extent, after debiasing, can we preserve model performance in both gender-related and other tasks?
    \item[C.] What are the similarities and differences in the distribution of two analyzed signals (bias and factual gender) in models' weights?
    \item[D.] Does \ddama{} generalize to other languages? Is the distribution of bias signal in a model shared across distinct languages?
\end{enumerate}

Upon publication of the work, we will release the code and instances of debiasd models.

\section{Methodology and Theoretical Background}
\label{sec:methodology}

In this section, we introduce \emph{Dual Debaising Algorithm through Model Adaptation} (\ddama{}), a new dual debiasing method.
We first describe the \ddama's intended function, i.e., how it should alter the pre-trained language model, and describe the algorithmic components that efficiently fulfill these aims. 
Then, we provide theoretical backing for the presented approach.
Appendix~\ref{sec:app-theory} contains the proofs and further terminological explanations.

\subsection{Motivations for 2DAMA}
\label{sec:2dama-aim}

The key idea of debiasing through model adaptation is that we can perform a targeted and efficient parameter change in a model to erase the unwanted signals encoded in it. 
The main risk of such an approach is that erasing bias can also affect the encoded knowledge.

We can draw a specific example of such a tradeoff from gender debiasing. 
We aim to reduce the models' reliance on stereotypes in predictions, e.g., 
given a stereotypical prompt as the one in Figure~\ref{fig:dama_prompt}: ``\textit{The salesperson laughed because}'', we intend to coerce equitable probabilities of possible gender predictions manifested by pronouns ``\textit{he}'', ``\textit{she}'', or ``\textit{them}''. 
On the contrary, when considering a prompt containing factual gender information: ``\textit{The king laughed because}'' the desired output distribution would assign a high probability to the male pronoun. 

We call the issue mentioned above \dd{} problem. 
\ddama{} is designed to alter language models to reduce their reliance on stereotypes while maintaining sensitivity to factual gender.

\subsection{Composing 2DAMA Approach}
\label{sec:2dama-composition}

In practice, \ddama{} leverage on the combination of new (\dd{}) and previously introduced algorithms (\dama{}, \leace{}). 
We shortly describe each of \ddama's components and the role they fulfill:

\paragraph{DAMA} \emph{Debiasing Algorithm through Model Adaptation} \cite{limisiewicz2024debiasing} is a method for adapting parameters of language models to mitigate the encoding of harmful biases without affecting their general performance.
The method employs model editing techniques \cite{meng_locating_2023} to disassociate specific signals provided in a prompt with the model outputs, i.e., stereotypes in prompts and gendered output.
In this work, we use most of \dama{} design choices, such as the adaptation of mid-upper feed-forward layers and algorithm to distill the stereotypical and gender-related parts of the latent representation (as depicted in Figure~\ref{fig:dama_schema}).

\paragraph{LEACE}  \emph{LEAst-squares Concept Erarsuer} \citep{belrose2023leace} is a method of concept erasure (such as bias signal) in latent representation. 
In Section~\ref{sec:dama-leace}, we show that combining \dama{} and \leace{} is possible and simplifies model adaptation. 
The simplification is obtained by replacing the Partial Least Squares concept erasure used in \dama{}, which required pre-defining the dimensionality of erased signals.
Moreover, in Section~\ref{sec:results-lm}, we show that \dama{} with \leace{} obtain comparable or better results.

\paragraph{2D} \dd{} is a new algorithm that we formally introduce in Section~\ref{sec:dd}.
The method uses covariance matrix decomposition to identify correlates related to bias and protected feature signals.
A concept erasure algorithm (based on \leace{}) is modified to erase bias while preserving protected features, in our case, factual gender.

\subsection{Model Adaptation with LEACE}
\label{sec:dama-leace}

\leace{} provides guarantees of erasing specific variable (concept $Z$) influence on another random variable (signal $X$). 
In the language of neural networks, we can consider latent vectors output by an intermediate layer as a sample drawn from the distribution given by $X$. \leace{} aims to de-correlate latent vectors with an unwanted signal (e.g., gender bias), whose distribution is represented as another random vector $Z$.

\begin{theorem}[\textbf{LEACE}]
\label{trm:leace}
We consider random vectors $X$  and $Z$ taking values in $\Real^n$. Both random vectors are centered, each with a finite moment.

Then the objective:
\begin{equation*}
    \argmin_{\mat{P} \in \Real^{n \times n}} \E\left[|| \mat{P}X -X||^2\right]
\end{equation*}
subject to:
\begin{equation*}
    \cov(\mat{P}X,Z)=0
\end{equation*}
is solved by:
\begin{equation*}
    \mat{P}^* = \Iden - \mat{W}^\dotplus \mat{P}_{\mat{W}\mat{\Sigma}}\mat{W},
\end{equation*}
where $\mat{W}$ is the whitening transformation $(\Sigma^{1/2}_{V,V})^\dotplus$;  $\mat{P}_{\mat{W}\mat{\Sigma}}$ is an orthogonal projection matrix onto colspace of $\mat{W}\mat{\Sigma}_{V,Z}$.
\footnote{Notation: $^\dotplus$ denotes Moonrose-Penrose psuedoinverse. For brevity, we use $\mat{\Sigma}_{V,Z}$ for covariance matrix $\cov(V,Z)$.
The complete terminological note can be found in Appendix~\ref{sec:app-theory}}
\end{theorem}

The proof can be found in \citet{belrose2023leace}. 
We extend this theoretical finding to erase the concept in linear transformations instead of latent vectors. 
We are specifically interested in transformation minimizing the error between input (keys: $U$) and predicted variable (values: $V$), as depicted in Figure~\ref{fig:dama_intervention}. 
We reasonably assume that dense layers of trained neural networks (e.g., feed-forward layers in Transformer) fulfill this purpose, i.e.:
\begin{equation}
    V = \mat{S}U - \epsilon,
\end{equation}
where $\mat{S}$ is a linear transformation and $\epsilon$ a vector of errors.
Due to gradient optimization in the model's pre-training, we assume that the feed-forward layer approximates the least solution, i.e., $FF \approx \mat{S}$.


Taking this assumption, we can present a theorem guaranteeing concept erasure in the model adaptation algorithm (\dama{}):

\begin{theorem}[\textbf{DAMA-LEACE}]
\label{trm:dama-leace}
We consider random vectors: $U$ taking values in $\Real^m$, $V$ and $Z$ taking values in $\Real^n$, where $m \geq n$. Under assumptions that: A) random vectors $U$, $V$, $Z$ are centered, and each of them has finite moment; B) the regression relation between $U$ and $V$ fulfill the assumption of ordinary least squares, and there exist least squares estimator $V = \mat{S}U - \epsilon$.

Then the objective:
\begin{equation*}
    \argmin_{\mat{P} \in \Real{n \times m}} \E\left[|| \mat{P}U -V||^2\right],
\end{equation*}
subject to:
\begin{equation*}
    \cov(\mat{P}U,Z)=0
\end{equation*}
is solved by:
\begin{equation*}
    \mat{P}^* = \left(\Iden - \mat{W}^\dotplus \mat{P}_{\mat{W}\mat{\Sigma}}\mat{W}\right)\mat{S},
\end{equation*}
where $\mat{W}$ is the whitening transformation $(\Sigma^{1/2}_{\mat{S}U,\mat{S}U})^\dotplus$;  $\mat{P}_{\mat{W}\mat{\Sigma}}$ is an orthogonal projection matrix onto colspace of $\mat{W}\mat{\Sigma}_{\mat{S}U,Z}$; $\mat{S}$ is a least squares estimator of $V$ given $U$:  $\mat{S}=\mat{\Sigma}_{U,V}\mat{\Sigma}_{U,U}^{-1}$.

\end{theorem}

Based on the theorem and the assumption that $FF \approx \mat{S}$ applying projections would break the correlation between stereotypical keys and gendered values with minimal impact on other correlations stored by the feed-forward layer.
We call the algorithm realizing such adaptation in a neural network: \dama-\leace.
\footnote{Noteworthy, \citet{belrose2023leace} also proposed using \leace{} to alter language model predictions. Yet they applied the algorithm to latent representation instead of altering model parameters as in \dama-\leace.}

\subsection{Dual Debiasing}
\label{sec:dd}

In \dd, we extend the concept erasure problem by considering two type signals and corresponding random variables: $Z_b$ bias to be erased and $Z_f$ feature to be preserved.
We posit that:

\begin{theorem}[\textbf{DUAL-DEBIASING}]
\label{trm:dual-debiasing}
We consider random vectors $X$, $Z_b$, and $Z_f$ in $\Real^n$. 
Under the assumptions that:
A)  $Z_b$ and $Z_f$ $Z_b \perp Z_f | X$, i.e., $Z_b$ and $Z_f$ are conditionally independent, given $X$;
B) $\mat{\Sigma}_{X,Z_b}\mat{\Sigma}_{X,Z_f}^T$, i.e., the variable $X$ is correlated with $Z_f$ and $Z_b$ through mutually orthogonal subspaces of $\Real^n$.
The solution of the objective:
\begin{equation*}
    \argmin_{\mat{P} \in \Real^{n \times n}} \E\left[|| \mat{P}X -X||^2\right],
\end{equation*}
subject to:
\begin{equation*}
    \cov(\mat{P}X,Z_b)=0,
\end{equation*}
satisfies:
\begin{equation*}
    \cov(\mat{P}X, Z_f)=\cov(X,Z_f).
\end{equation*}
\end{theorem}

The theorem shows that the correlation with the conditionally independent features is left intact by applying \leace{} erasure to a bias signal.
However, the assumption of conditional independence is strong and unlikely to hold when considering the actual signals encoded in the model.
Thus, for practical applications, we need to relax the requirements.

In \dd{}, we abandon theoretical guarantees of the theorem to consider bias and feature signals that can be conditionally correlated.
In constructing the debiasing projection ($\mat{P}*$), we must decide whether specific dimensions should be nullified or preserved.
We propose to nullify dimensions of $X$  with $t$ times higher correlation with $Z_f$ than $Z_b$, where the threshold $t$ (later referred to as \emph{bias-to-feature threshold}) is empirically chosen.
To analyze the correlations we consider correlation matrix $\mat{W}\mat{\Sigma}_{X,[Z_f,Z_b]}$.
By using singular value decomposition, we can identify the share of variance in each column's first $n$ rows (associated with $Z_f$) and the latter $n$ rows (associated with $Z_b$). 
In modified colspace projection $\mat{\tilde{P}}_{\mat{W}\mat{\Sigma}}$, we only consider the column with  $t$ times higher variance with $Z_f$ than with $Z_b$.
Thus the final \dd{} \leace{} projection $\mat{\tilde{P}}^* = \left(\Iden - \mat{W}^\dotplus \mat{\tilde{P}}_{\mat{W}\mat{\Sigma}}\mat{W}\right)$ will to large extent preserve the protected feature while reliably erasing bias. 
In Section~\ref{sec:bias-to-feature}, we study the impact of feature-to-bias threshold $t$ in practice.

\section{Experimental Setting}
\label{sec:exp-setting}

\begin{figure*}[t]
     \centering
     \begin{subfigure}[b]{0.75\textwidth}
         \centering
        \begin{subfigure}[t]{\textwidth}
             \centering
             $\text{EN: ``The \textcolor{schema_celadon}{\textbf{salesperson}} laughed because \{ he | she \}''}$
             $\text{EN: ``The \textcolor{schema_celadon}{\textbf{saleseperson}} is not working today.''} \rightarrow \text{DE: \{ Der | Die \}''}$ 
              $\text{EN: ``That \textcolor{schema_celadon}{\textbf{saleseperson}} is not working today.''} \rightarrow \text{CS: \{ Ten | Ta \}''}$
             
             \caption{}
             \label{fig:dama_prompt}

        \end{subfigure}
        \begin{subfigure}[b]{0.52\textwidth}
         \includegraphics[width=\textwidth]{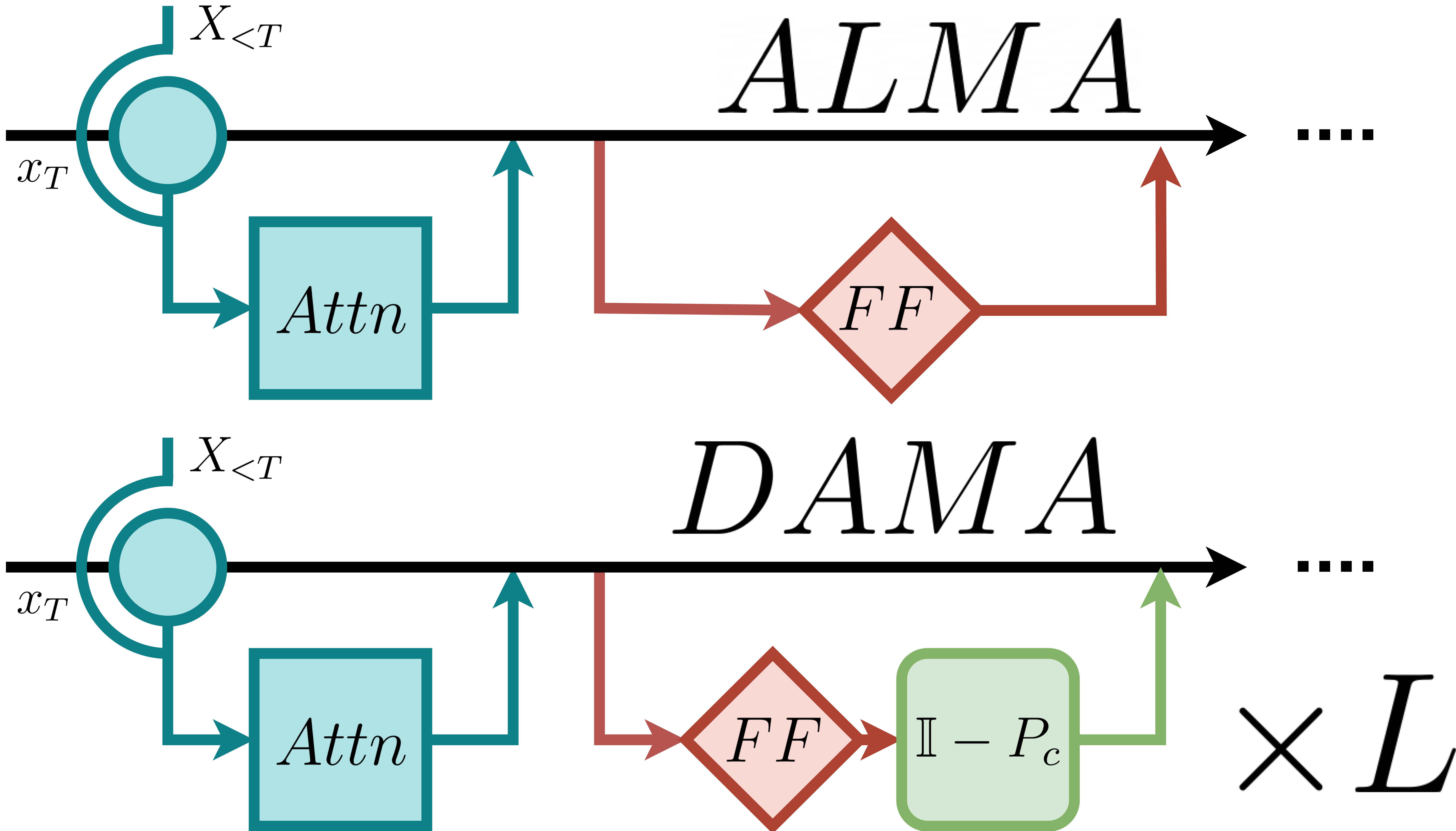}
         \caption{}
         \label{fig:dama_intervention}
        \end{subfigure}
        \hfill
        \begin{subfigure}[b]{0.47\textwidth}
         \centering
         \includegraphics[width=\textwidth]{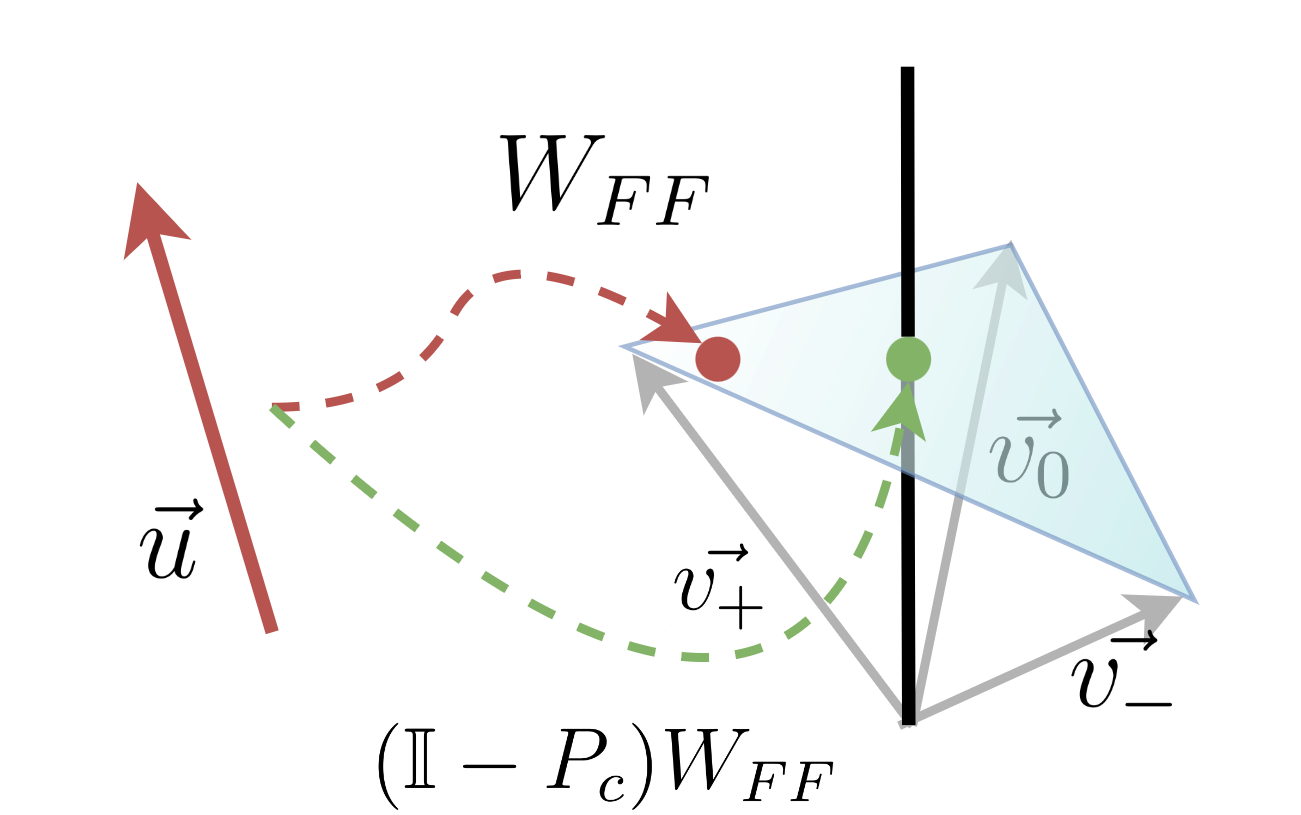}
         \caption{}
         \label{fig:dama_projection}
        \end{subfigure}
    \end{subfigure}
    \hfill
    \begin{subfigure}[b]{0.24\textwidth}
         \centering
         \includegraphics[width=\textwidth]{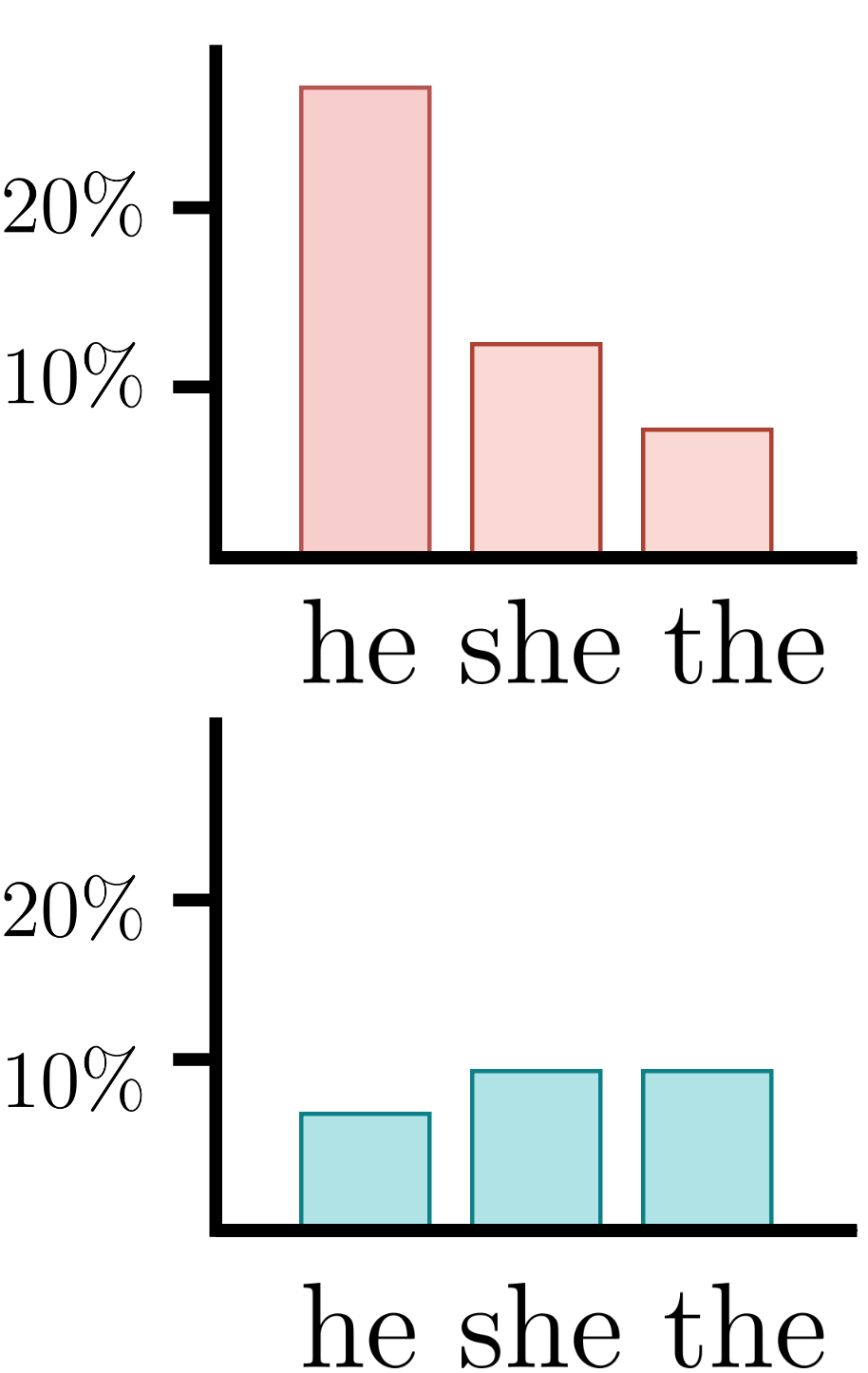}
         \caption{}
         \label{fig:dama_probabilities}
     \end{subfigure}

     \caption{Schema (b) shows \dama{} intervention in a language model layer.
    (a) We show the model's behavior when presented with a stereotypical prompt in three languages.
    Specifically, (c) shows the projections of the feed-forward latent vector ($\vec{u}$) onto the output space. With \dama{} (\textcolor{schema_green}{lower arrow}), we nullify the gender component of the representation. It results in balanced probabilities of gendered tokens in the model's output, as shown in (d). Adapted from \citet{limisiewicz2024debiasing}.}
     \label{fig:dama_schema}
\end{figure*}

This section presents an empirical setting to examine the practical application of model editing methods.
Specifically, we describe models, data, and evaluation metrics for gender bias and general performance.

\subsection{Models}

In experiments, we focus on \llama{} family models \cite{touvron_llama2_2023,dubey2024llama}, which are robust and publically available language models developed by Meta AI.
We analyze \llama{}~2 models of sizes 7 and 13 billion parameters and \llama{}~3 with 8 billion parameters.
In multilingual experiments, we use ALMA-R 13 billion parameter model \cite{xu_contrastive_2024}.
ALMA-R is based on an instance of the Llama 2 model that was fine-tuned to translate using Contrastive Preference Optimization.
ALMA-R covers translation between English and five languages (German, Czech, Russian, Icelandic, and Chinese) and shows competitive translation quality results.

In model editing experiments, we adapt the layers starting from the one found in the two-thirds of the layer stack counted from the input to the output. 
It is the 26th layer for 13 billion parameter models and the 21st layer for smaller models.
For example, the adaptation of 11 mid-upper layers in the 13B model modifies the layers from 26th through 37th.


\subsection{Data for Dual Debiasing}
\label{sec:data-for-debiasing}
Following \citet{limisiewicz2024debiasing}, we feed prompts to the model in order to obtain the latent embeddings in the input of latent layers. We treat these embeddings as key vectors ($U$) containing stereotypical or factual gender signals. To obtain the gendered value vectors ($V$), we find the layer's output vector that would maximize the probability of predicting tokens corresponding to gender.

\paragraph{Language Modeling Prompts}
For debiasing language models, we use solely English prompts. 
We design 11 prompt templates, such as
``\textit{The \textcolor{schema_celadon}{\textbf{X}} laughed because \_\_\_}'', where ``\textcolor{schema_celadon}{\textbf{X}}'' should be replaced by profession name.
This prompt construction provokes the model to predict one of the gendered pronouns (``\textit{he}'', ``\textit{she}'', or ``\textit{them}'').
To distinguish stereotypical signals for debiasing, we use 219 professions without factual gender that were annotated as stereotypically associated with one of the genders by \citet{bolukbasi_2016}. 

\paragraph{Multilingual Prompts}
For debiasing machine translation, we use prompts instructing the model to translate sentences containing the same set of 219 professions to a target language that has the grammatical marking of gender, e.g.,  ``\textit{English: The \textcolor{schema_celadon}{\textbf{X}} is there. German: \_\_\_}''.
The translation model would naturally predict one of the German determiners, which denotes gender (``\textit{Der}'' for male or ``\textit{Die}'' for female).
For each model, we adjust the template to include instructions suggested by the model's authors.
We construct the translation prompts for two target languages, Czech and German, proposing 11 templates for each.

\paragraph{Factual Prompts}
\emph{Dual debiasing} requires using factual prompts to identify the signal to be preserved.
For that purpose, we use the same prompt templates as defined above (both English and multilingual) with the distinction of entities used to populate them.
For that purpose, we propose 13 pairs of factually male and female entities, e.g., ``\textit{king}'' -- ``\textit{queen}'', ``\textit{chairman}'' -- ``\textit{chairwoman}''.

The examples of language modeling and multilingual prompts are given in Figure~\ref{fig:dama_prompt}.
We list all of the prompt templates in Appendix~\ref{sec:app-prompts}

\subsection{Bias Evaluation}
\label{sec:bias-evaluation}
\paragraph{Language Modeling}
We assess the bias in language generation following the methodology of \citet{limisiewicz2024debiasing}.
From the dataset of \citet{bolukbasi_2016}, we select the held-out set of professions that were not included in the 219 used for debiasing.
Each profession was assigned two scores: \fact{} score $x_f$ and \stereo{} score $x_s$. 
The scores define how strongly a word (or a prompt) is connected with the male or female gender, respectively, through factual or stereotypical cues. 
By convention, scores range from $-1$ for female-associated words to $1$ for male ones.
We measure the probabilities for gendered prediction for a given prompt $P_M(o|X)$.
For that purpose, we use pronouns $o_+ = \text{``he''}$ and $o_- = \text{``she''}$, as they are probable continuations for given prompts.
Subsequently for each prompt, we compute \empiric{} score $y = P_M(o_+|X) - P_M(o_-|X)$. 
To estimate the relationship between the observed score and annotated ones $x_s$ and $x_f$, we construct a linear model:
\begin{equation}
\label{eq:linear_model}
    y = a_s \cdot x_s + a_f \cdot x_f + b_0
\end{equation}
The linear fit coefficients have the following interpretations: $a_s$ is an impact of stereotypical signal on the model's predictions; 
$a_f$ is an impact of the factual gender of the word. 
Noticeably, $y$, $x_s$, and $x_f$ take the values in the same range.
The slope coefficient tells how shifts in annotated scores across professions impact the difference in prediction probabilities of male and female pronouns.
The intercept $b_0$  measures how much more probable the male pronouns are than the female pronouns when we marginalize the subject.

\begin{table}[t!]
\centering
\begin{adjustbox}{max width=\linewidth}
\footnotesize
\begin{tabular}{@{}lccccc@{}}
\toprule
           & \multicolumn{3}{c}{Bias in LM}                           & \multicolumn{2}{c}{WinoBias}                      \\ \cmidrule(rl){2-4} \cmidrule(l){5-6}
           & $\downarrow$ $a_s$ & $\uparrow$ $a_f$ & $\downarrow$ $b$ & $\downarrow$ $\Delta S$ & $\downarrow$ $\Delta G$ \\ \midrule
\llama{}~2 7B  & 0.234 & 0.311 & 0.090 & 33.6 & 7.3 \\
DAMA       & 0.144 & 0.205 & 0.032 & 27.3 & 6.8 \\
DAMA+LEACE & 0.118 & 0.171 & 0.028 & 22.9 & 5.4 \\
2DAMA      & 0.128 & 0.187 & 0.042 & 22.9 & 5.7 \\ \midrule
\llama{}~2 13B & 0.244 & 0.322 & 0.097 & 35.0 & 0.3 \\
DAMA       & 0.099 & 0.160 & 0.030 & 26.4 & 2.4 \\
DAMA+LEACE & 0.098 & 0.159 & 0.026 & 26.5 & 2.4 \\
2DAMA      & 0.119 & 0.206 & 0.023 & 27.0 & 1.9 \\ \midrule
\llama{}~3 8B  & 0.262 & 0.333 & 0.082 & 36.8 & 2.7 \\
DAMA       & 0.069 & 0.090 & 0.144 & 20.3 & 4.2 \\
DAMA+LEACE & 0.084 & 0.157 & 0.082 & 18.8 & 2.7 \\
2DAMA      & 0.140 & 0.209 & 0.051 & 18.7 & 2.4 \\ \bottomrule
\end{tabular}
\end{adjustbox}
\caption{Bias evaluation for the \llama{}~ family models, and their adaptation with different debiasing algorithms (\dama{}, \dama{} with \leace{}, and \ddama{}).
The debiasing adaptation was applied to 12  mid-upper layers for the 13B model and 9 mid-upper layers for the smaller ones.
In \ddama{}, we set bias-to-feature threshold to $t=0.05$.
}
\label{tab:2dama-bias}
\end{table}

\paragraph{Other Bias Manifestations in English}
We evaluate the bias in coreference resolution based on \textbf{WinoBias} dataset \citep{zhao-etal-2018-gender}. 
We use metrics $\Delta G$ and $\Delta S$ to evaluate representational and stereotypical bias, respectively. 
$\Delta G$ measures the difference in coreference identification correctness (accuracy) between masculine and feminine entities; similarly, $\Delta S$ measures the difference in accuracy between pro-stereotypical and anti-stereotypical instances of gender role assignments. 

\begin{table}[t]
\centering
\footnotesize
\begin{tabular}{@{}lccc@{}}
\toprule
           & LM               & \multicolumn{2}{c}{ARC}                 \\ \cmidrule(lr){2-2} \cmidrule(l){3-4} 
           & $\downarrow$ ppl & $\uparrow$ acc (C) & $\uparrow$ acc (E) \\ \midrule
\llama{}~2 7B  & 21.28            & 70.2             & 42.5             \\
DAMA       & 21.51            & 69.8             & 42.8             \\
DAMA+LEACE & 23.81            & 68.3             & 41.2             \\
2DAMA      & 23.66            & 67.5             & 42.0             \\ \midrule
\llama{}~2 13B & 19.68            & 72.6             & 46.8             \\
DAMA       & 18.94            & 71.6             & 45.0             \\
DAMA+LEACE & 19.67            & 71.3             & 46.4             \\
2DAMA      & 19.90            & 71.2             & 46.1             \\ \midrule
\llama{}~3 8B  & -                & 67.1            & 39.9            \\
DAMA       & -                & 64.6            & 38.1            \\
DAMA+LEACE & -                & 63.0            & 39.8            \\
2DAMA      & -                & 63.5            & 37.9            \\ \bottomrule
\end{tabular}
\caption{General performance in language modeling and reasoning on ARC \textbf{C}halange and \textbf{E}asy subset. We present results for \llama{} family models, and their adaptation with different debiasing algorithms (\dama{}, \dama{} with \leace{}, and \ddama{}).
We do not present perplexity for \llama{} 3 because the model has different vocabulary and the results are not comparable.
The hyperparameters are the same as in Table~\ref{tab:2dama-bias}}
\label{tab:2dama-general}
\end{table}

\paragraph{Translation}
\citep{stanovsky-etal-2019-evaluating} proposed using Winograd Challenge sentences
for evaluating bias in translation from English into eight languages
with morphological marking of gender (e.g., German, Spanish, Russian, Hebrew). 
In \textbf{WinoMT}, the correctness of the translation is computed by the F1 score of correctly generating gender inflection of profession words in the target language.
The evaluation of gender bias is analogical, as in WinoBias.
$\Delta G$ and $\Delta S$ measure the difference in F1 scores: male vs. female and pro- vs. anti-stereotypical sets of professions, respectively. 
The more recent \textbf{BUG} \cite{levy-etal-2021-collecting-large} dataset is based on the same principle of bias evaluation, with the distinction that it contains naturally occurring sentences instead of generic templates used in WinoMT.


\subsection{General Performance Evaluation}

\paragraph{Language Modeling}
 We evaluate perplexity on general domain texts from \textbf{Wikipedia-103} corpus \citep{merity2016pointer}. 

\paragraph{Reasoning Endtask}
To assess the models' reasoning capabilities, we compute accuracy on \textbf{AI2 Reasoning Challenge (ARC)} \citep{clark2018think} in both easy and challenging subsets. 

\paragraph{Translation}
To monitor the effect of debiasing on translation quality, we evaluate models on \textbf{WMT-22} \citep{kocmi-etal-2020-gender} 
parallel corpora with German, Czech, and Russian sentences and their translations in English.
We estimate the quality by two automatic metrics: COMET-22 \cite{rei-etal-2022-comet} and chrf \cite{popovic-2015-chrf}.

\begin{figure}[t!]
    \centering
    \includegraphics[width=\linewidth]{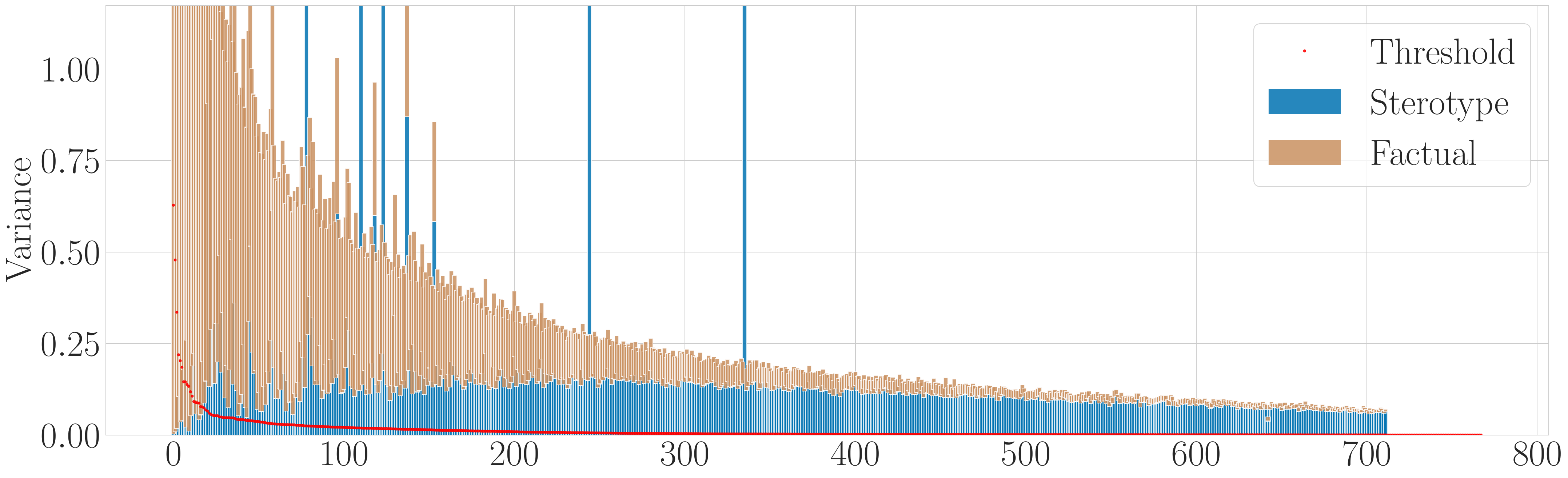}
    \caption{Visualization of dimensions and their variances related to stereotypical and factual gender signals identified by \dd{} algorithm in 26th layer of \llama{}~2 13B.
    The red dots denote the bias-to-feature threshold $t=0.05$.
    In \ddama{}, the dimension is preserved if stereotypical covariance is below the threshold.}
    \label{fig:concept-erasure-26L}
\end{figure}

\section{Debiasing Language Models}
\label{sec:results-lm}

In the first batch of the experiments, we evaluate the effectiveness of debiasing language models with model editing.
In these experiments, we solely focus on tasks in English.
We specifically analyze three model editing approaches: \dama{} as a baseline; \dama{} in combination with \leace{}; and \ddama{}, which employs \dd{} to preserve factual gender information.

\subsection{Main Results}

\paragraph{Model editing reduces bias and preserves the model's performance.}
All of the considered methods reduce gender bias both in language modeling and coreference resolution (Table~\ref{tab:2dama-bias}).
Remarkably, we observe that the model's overall performance, i.e., unrelated to gender, is not significantly affected, as demonstrated by perplexity and question-answering results (Table~\ref{tab:2dama-general}).
Relatively worse performance preservation was observed for \llama{}~3, which could be caused by intervening in too many layers.

\begin{table}[t!]
\centering
\small
\begin{tabular}{@{}ccccc@{}}
\toprule
\multirow{2}{*}{Layer} & \multicolumn{2}{c}{Bias Dimesnions} & \multicolumn{2}{c}{Variance Erased} \\ \cmidrule(rl){2-3} \cmidrule(l){4-5}
   & Erased & Preserved & Bias   & Factual \\ \midrule
26 & 712    & 12        & 99.6\% & 69.4\%  \\
27 & 774    & 18        & 99.4\% & 64.0\%  \\
28 & 782    & 22        & 99.0\% & 62.4\%  \\
29 & 750    & 17        & 99.5\% & 65.4\%  \\
30 & 713    & 19        & 99.5\% & 64.0\%  \\
31 & 304    & 12        & 99.3\% & 57.6\%  \\
32 & 387    & 16        & 99.2\% & 57.0\%  \\
33 & 469    & 17        & 99.2\% & 60.1\%  \\
34 & 716    & 21        & 99.2\% & 61.3\%  \\
35 & 621    & 18        & 99.2\% & 62.2\%  \\
36 & 406    & 20        & 98.9\% & 54.0\%  \\
37 & 409    & 18        & 99.1\% & 57.2\%  \\ \bottomrule
\end{tabular}
\caption{Number of erased and preserved orthogonal dimensions with \ddama{} in each feed-forward layer.
We call a dimension ``\textit{biased}'' when it belongs to col-space spanned by covariance matrix between latent representation and bias signal ($\mat{W}\mat{\Sigma}_{\mat{S}U,Z}$).
We present the percentage of erased covariance with stereotypical bias and factual gender as the result of the intervention in the layers. 
The bias-to-feature threshold was set at $t=0.05$.}
\label{tab:concept-erasure-dims-var-0.05}
\end{table}

\paragraph{Streamlining the approach with \leace{}.}
We observe that \dama{}-\leace{} reduces bias to a larger extent than baseline  \dama{}. 
The more substantial debiasing effect comes in pair with a slightly higher drop in general performance, as shown in Table~\ref{tab:2dama-general}. 
Yet, the deterioration is still insignificant compared to the original models' scores.
The crucial benefit of \dama{}-\leace{} is that projection dimensionality does not need to be pre-defined because it is learned implicitly (details in Section~\ref{sec:dama-leace}).
\footnote{In baseline \dama{}, the projection dimensionality is pre-set to $d=256$ for the 7B model and $d=512$ for the 13B models.}
That motivates us to use \dama{}-\leace{} in further experiments.

\begin{figure*}[!tb]
     \centering
     \begin{subfigure}[b]{0.4\textwidth}
         \centering
         \includegraphics[width=\textwidth]{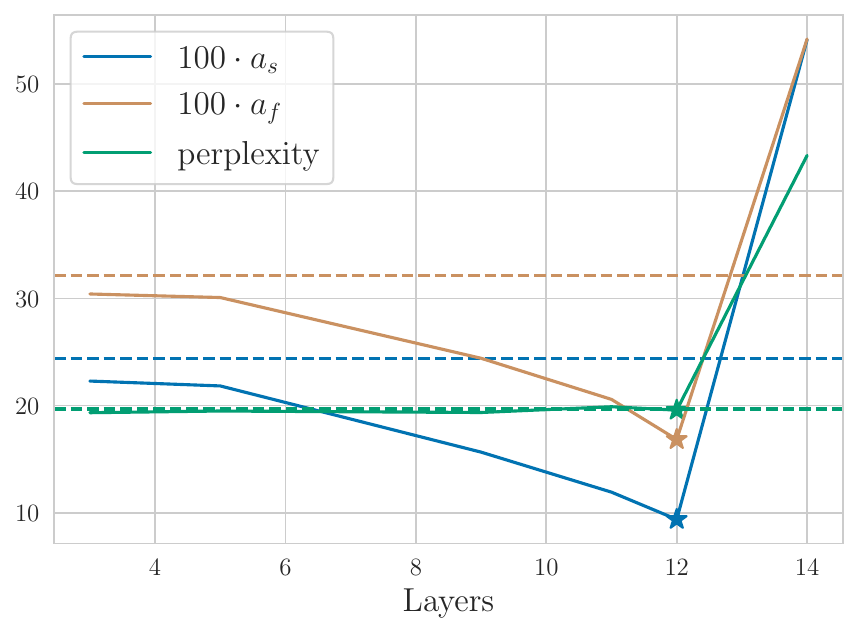}
         \caption{Bias-to-feature threshold fixed at 0.05}
     \end{subfigure}
     \begin{subfigure}[b]{0.4\textwidth}
         \centering
         \includegraphics[width=\textwidth]{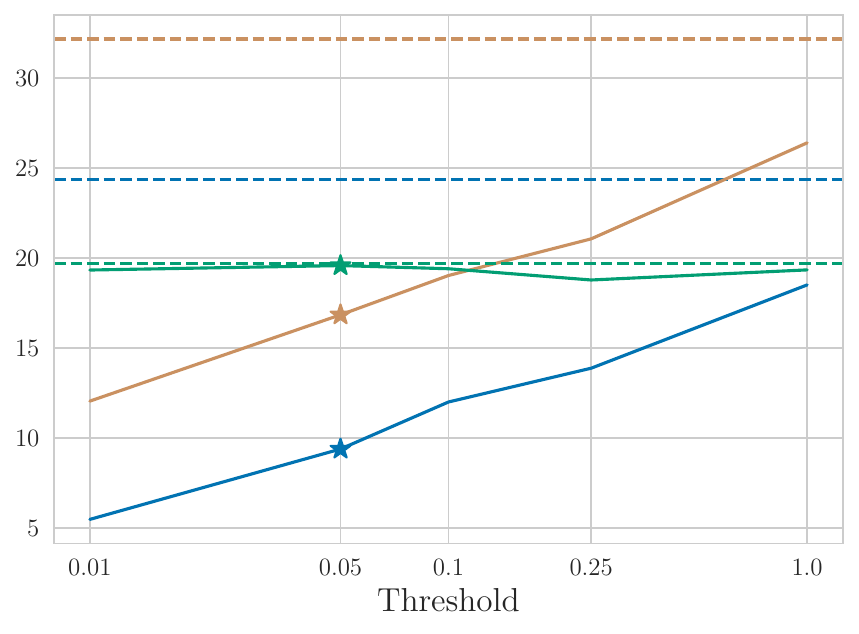}
         \caption{Number of layers fixed at 12}
     \end{subfigure}
     
        \caption{The hyperparameter analysis for \ddama{} applied to \llama{}~2 13B model on performance and bias in language modeling. We measured bias on gendered prompts by linear coefficients: $a_s$ and $a_f$, the language modeling capabilities are measured by perplexity. Stars mark the performance of the best setting.
        The dashed line corresponds to the scores of the original model.}
        \label{fig:abaltion_llama2}
\end{figure*}

\paragraph{Preserving factual gender with \dd{}.}
The coefficients $a_s$ and $a_f$ from Table~\ref{tab:2dama-bias} indicate how much the models' prediction is affected by gender present through stereotypical and factual cues, respectively. 
We see that \ddama{} enables, to a significant extent, preserving factual gender information with a slight increase in susceptibility to gender bias.

In the remainder of this Section, we describe the relationship between stereotype and factual gender encoding in the model and the role of parameter choice in \ddama{}.

 \subsection{Relationship between Seterotypical and Factual Signals}
\label{sec:bias-to-feature}

With \dd{}, we can analyze the covariance of the latent orthogonal dimensions in the model's feed-forward layers with the stereotypical and factual signals (as detailed in Section~\ref{sec:dd}).
In Figure~\ref{fig:concept-erasure-26L}, we plot these covariances for each dimension.
The visualization reveals that the factual gender is represented by relatively few dimensions with high covariance.
In contrast, stereotypical bias is encoded in more-dimensional subspaces, yet each dimension has low covariance.

In Table~\ref{tab:concept-erasure-dims-var-0.05}, we demonstrate that in \ddama{} with low bias-to-feature threshold $t=0.05$ preserves only a few dimensions responsible for stereotypical bias in each layer.
Such intervention in the model erases $\approx99\%$ of covariance with a stereotypical signal while keeping over $30\%$ of covariance with a factual gender signal.

\subsection{Choice of Hyperparameters}
\label{sec:hyperparameters}

We present the impact of two parameters on the effectiveness of \ddama{} in Figure~\ref{fig:abaltion_llama2}.
The first is the bias-to-feature threshold $t$.
We observe that its choice controls the trade-off between mitigating bias and preserving factual information.
We set it a low value of $0.05$ because our primary objective is the reduction of bias.
The second hyperparameter is the number of layers that should be edited.
We confirm the findings of \citet{limisiewicz2024debiasing} that adaptation should applied to approximately one-third of the midd-upper layers.
Notably, the top two layers (38th and 39th) should be left out.

\section{Beyond English: Multilingual Debiasing}
\label{sec:results-mt}

In a multilingual setting, we
debias a model fine-tuned for translation: ALMA-R 13B \cite{xu_contrastive_2024} by employing the collection of the new multilingual debiasing prompts.
We specifically focus on gender bias and quality of translation between English and Czech, German, and Russian.

\subsection{Main Results}

\paragraph{Model editing generalizes to the multilingual settings.}
Analogically to experiments in English, we show that model editing reduces bias in translation and has a mediocre impact on the translation quality (as shown in Table~\ref{tab:2dama-translation}).
We observe some differences in results between the two analyzed languages. 
Overall, the scores after debiasing are better for German than Czech, indicating that German prompts are better quality.

\begin{table*}[!th]
\centering
\begin{adjustbox}{max width=\linewidth}
\small
\begin{tabular}{@{}llcccccccc@{}}
\toprule
\multirow{2}{*}{Language} & \multirow{2}{*}{} & \multicolumn{2}{c}{Translation to English} & \multicolumn{2}{c}{Translation from English} & \multicolumn{2}{c}{WinoMT} & \multicolumn{2}{c}{BUG} \\ \cmidrule(r){3-4} \cmidrule(rl){5-6} \cmidrule(rl){7-8} \cmidrule(l){9-10}   
 &  & $\uparrow$ comet & $\uparrow$ chrf & $\uparrow$ comet & $\uparrow$ chrf & $\downarrow$ $\Delta S$ & $\downarrow$ $\Delta G$ & $\downarrow$ $\Delta S$ & $\downarrow$ $\Delta G$ \\ \midrule
\multirow{4}{*}{German} & ALMA-R 13B & 85.0 & 57.0 & 86.7 & 58.1 & 30.5 & 3.7 & 7.8 & 32.5 \\
 & DAMA+LEACE & 85.0 & 56.7 & 85.3 & 55.4 & 20.5 & 10.0 & 5.4 & 33.6 \\
 & 2DAMA ($t=0.05$) & 84.9 & 56.7 & 85.1 & 54.8 & 22.6 & 3.3 & 4.4 & 27.8 \\
 & 2DAMA ($t=1.00$) & 84.9 & 56.6 & 85.4 & 55.4 & 22.1 & -10.1 & 7.7 & 28.4 \\ \midrule
\multirow{4}{*}{Czech} & ALMA-R 13B & 87.0 & 68.6 & 89.7 & 53.8 & 26.3 & 2.1 & 11.7 & 9.2 \\
 & DAMA+LEACE & 86.9 & 68.2 & 88.6 & 50.1 & 21.6 & 17.7 & 10.4 & 18.0 \\
 & 2DAMA ($t=0.05$) & 86.9 & 68.1 & 88.5 & 49.9 & 18.0 & 14.6 & 4.5 & 11.0 \\
 & 2DAMA ($t=1.00$) & 86.9 & 68.1 & 88.8 & 50.4 & 22.4 & 7.2 & 8.6 & 9.8 \\ \bottomrule
\end{tabular}
\end{adjustbox}
\caption{Evaluation of gender bias and quality of translation.
In all the methods, ALMA-R was used as the base model.
Adaptations were applied to 11 mid-upper feed-forward layers. 
Translation quality was evaluated on the WMT-22 dataset. 
}
\label{tab:2dama-translation}
\end{table*}
\begin{table*}[!ht]
\centering
\begin{subtable}[c]{0.45\textwidth}
\small
\begin{tabular}{@{}cccc@{}}
\toprule
Prompt Lang. $\downarrow$ & German & Czech & Russian \\ \midrule
$\emptyset$   & 3.7    & 2.1   & 25.7    \\
English       & 11.1   & 7.9   & 31.4    \\
German        & 3.3    & 21.6  & 31.3    \\
Czech         & 6.2    & 14.6  & 32.0     \\
All Above     & 8.1    & 23.2  & 33.4    \\ \bottomrule
\end{tabular}
\caption{Representational bias ($\Delta G$)}
\label{tab:xlingual-deltaG}
\end{subtable}
\hspace{0.5cm}
\begin{subtable}[c]{0.45\textwidth}
\small
\begin{tabular}{@{}cccc@{}}
\toprule
Prompt Lang. $\downarrow$ & German & Czech & Russian \\ \midrule
$\emptyset$    & 30.5   & 26.3  & 10.2    \\
English        & 28.5   & 21.2  & 7.0     \\
German         & 14.4   & 15.1  & 4.0     \\
Czech          & 24.3   & 17.2  & 3.9     \\
All Above      & 24.0   & 18.7  & 1.3     \\ \bottomrule
\end{tabular}
\caption{Stereotypical bias ($\Delta S$)}
\label{tab:xlingual-deltaS-12l}
\end{subtable}
\caption{Bias evaluation based on WinoMT challenge-set. The evaluation language is shown at the top of each column.
Each row corresponds to a set of languages for which prompts were used in model adaptation.
The debiasing adaptation was performed with \ddama{} on 11 mid-upper layers with the bias-to-feature threshold set to $t=0.05$.
}
\end{table*}

\paragraph{\dd{} is required to mitigate representational bias.}
Our methods are more effective for the stereotypical manifestation of bias $\Delta S$ than the representational one $\Delta G$.
In the representational bias, we sometimes observe bias increase after model editing.
To remedy that, we use \ddama{} with higher values of feature-to-bias threshold ($t=1.00$ instead of $t=0.05$), which tends to preserve more factual signal.
Factual gender understanding is especially essential for equitable representation of factual gender in morphologically rich languages, as evidenced by $\Delta G$ scores for $t=1.00$ setting.
This finding emphasizes the utility of \ddama{} in a multilingual setting. 

The extended study of hyperparameters in translation debiasing is presented in Appendix~\ref{sec:app-results}.

\subsection{Cross-lingual Debiasing}

An intriguing question of multilingual bias is whether its encoding is shared across languages \cite{gonen-etal-2022-analyzing}.
We test this hypothesis by editing models with prompts in one or multiple languages and testing on another language.
The results show evidence of effectiveness in cross-lingual mitigation of stereotypical gender bias.
In Table~\ref{tab:xlingual-deltaS-12l}, we observe that some languages are more effective in debiasing than others, e.g., German prompts offer the strongest $\Delta S$ reduction for both Czech and German. 
Whereas to control representational bias mitigation ($\Delta G$), it is recommended to use in-language prompts, as indicated by Czech, German, and Russian results in Table~\ref{tab:xlingual-deltaG}.




\section{Related Work}

We refer to the related work grouped into three areas.

\subsection{Model Editing and Concept Erasure}

Model editing is a method of applying targeted changes to the parameters of the models to modify information encoded in them.
Notable examples of model editing include targeted changes in the model's weight \cite{mitchell2022fast, meng_mass-editing_2022, meng_locating_2023} or adaptation with added modules (adapters) \cite{houlsby2019adapters, hu2022lora}.
The technique showed promising results as the tool to erase specific information \cite{patil2024can}.

In the literature, bias mitigation was perceived as a theoretically interesting and practical application for concept erasure.
\citet{ravfogel-etal-2020-null, ravfogel_linear_2022, belrose2023leace} proposed effective linear methods of erasing gender bias from the latent representation of language models.
Other approaches aimed to edit pre-trained language models to reduce their reliance on stereotypes.
They include: causal intervention \citep{vig_causal_2020}, model adapters \citep{fu_adapterbias_2023}, or targeted weight editing \cite{limisiewicz2024debiasing}.
As the name indicates, \ddama{} primarily builds on the last method.

\subsection{Debiasing Machine Translation}

Machine translation systems have been shown to exhibit gender bias in their predictions  \cite{savoldi_etal_2021}.
The problem is especially severe in translation from languages that do not grammatically mark gender (e.g., English, Finish) to ones that do (e.g., German, Czech, Spanish) because translation requires predicting gender, which is not indicated in the reference \cite{stanovsky-etal-2019-evaluating}.
There have been a few past attempts to mitigate biases in translation systems \citep{saunders-byrne-2020-reducing, iluz-etal-2023-exploring, zmigrod-etal-2019-counterfactual}.
Nevertheless, these approaches are based on fine-tuning for non-stereotypical sentences, which increases the model's specialization but significantly reduces usability, e.g., in tasks unrelated to gender \citep{Luo2023AnES}. 

One key constraint of multilingual debiasing is the scarcity of bias annotations in various languages.
Notable datasets were introduced by \citet{levy-etal-2021-collecting-large, neveol-etal-2022-french}.
The difficulty of obtaining reliable cross-lingual bias resources stems from the need for deep knowledge of culture in addition to understanding a language. 
For instance, in different cultures, the same mentions tend to be associated with distinct sets of stereotypes:
the word ``\textit{doctor}'' in English is strongly associated with the male gender due to the overrepresentation of men in this profession in the US and the UK.
While in many Slavic and Baltic languages, the ``\textit{doctor}''s are to a lesser extent biased or even associated with the female gender, due to the higher proportion of women in medical professions.\footnote{According to \citet{oecd-helth-glance-2023}: in 2021, women made up over 70\% of medical doctor numbers in Lithuania, Latvia, and Estonia, while only 36\% in the US .}

To the best of our knowledge, we are the first to propose a method for debiasing LLM in machine translation tasks.

\subsection{Controlled Debiasing}

Sound practice in model debiasing involves carefully monitoring performance in bias-unrelated tasks and diverse manifestations of bias.
The former aspect is needed for the reliability and usefulness of the models after debiasing.
The latter's importance is highlighted by the observations that different bias metrics are mutually weakly or even negatively correlated \citep{delobelle-etal-2022-measuring, vanderwal2023undesirable}, raising the need for holistic evaluation.
The aspect of preserving useful features correlated with bias is especially underresearched.
A noteworthy study of interactions between factual gender and gender bias was conducted by \citet{limisiewicz-marecek-2022-dont}.
To our knowledge, we are the first to show that keeping factual gender can help mitigate diverse types of bias.

\section{Disussion and Conclusion}

We highlight the importance of considering the dual character of gender encoding in model editing.
Our theoretical and empirical results show that the novel model editing methods: \ddama{} effectively reduces the impact of stereotypical bias on the predictions while preserving equitable representation of (factual) gender based on grammar and semantics.
Maintaining the factual component of gender representation is crucial for debiasing in languages other than English, which denote gender more ubiquitously. 
Furthermore, our method does not significantly deteriorate the high performance of LLMs in various evaluation settings unrelated to gender.

To answer the research questions.
The results show that novel \ddama{} is an effective debiasing method that can implicitly learn the extent (dimensionality) of bias signals (re: A). 
Thus, it is easier to apply to new models.
Specifically, we do not require setting the number of erased bias dimensions in latent space thanks to changing the underlying erasure to \leace{}.
Furthermore, our method has minimal impact on gender-unrelated tasks (re: B), and thanks to \dd{}, it can also identify and preserve factual signals, even though it is significantly correlated with bias (re: C).
Lastly, we show that \ddama{} can be applied to a multilingual setting, and we observe the evidence of bias sharing across languages (re: D).

\section*{Acknowledgements}
We have been supported by the grant 23-06912S of the Czech Science Foundation. We have been using language resources and tools developed, stored, and distributed by the LINDAT/CLARIAH-CZ project of the Ministry of Education, Youth and Sports of the Czech Republic (project LM2018101).

\bibliography{custom, anthology-trimmed}

\begin{thebibliography}{42}
\expandafter\ifx\csname natexlab\endcsname\relax\def\natexlab#1{#1}\fi

\bibitem[{Belrose et~al.(2023)Belrose, Schneider-Joseph, Ravfogel, Cotterell, Raff, and Biderman}]{belrose2023leace}
Nora Belrose, David Schneider-Joseph, Shauli Ravfogel, Ryan Cotterell, Edward Raff, and Stella Biderman. 2023.
\newblock \href {https://openreview.net/forum?id=awIpKpwTwF} {{LEACE}: Perfect linear concept erasure in closed form}.
\newblock In \emph{Thirty-seventh Conference on Neural Information Processing Systems}.

\bibitem[{Bolukbasi et~al.(2016)Bolukbasi, Chang, Zou, Saligrama, and Kalai}]{bolukbasi_2016}
Tolga Bolukbasi, Kai{-}Wei Chang, James~Y. Zou, Venkatesh Saligrama, and Adam~Tauman Kalai. 2016.
\newblock Man is to computer programmer as woman is to homemaker? debiasing word embeddings.
\newblock In \emph{Advances in Neural Information Processing Systems 29: Annual Conference on Neural Information Processing Systems 2016, December 5-10, 2016, Barcelona, Spain}, pages 4349--4357.

\bibitem[{Clark et~al.(2018)Clark, Cowhey, Etzioni, Khot, Sabharwal, Schoenick, and Tafjord}]{clark2018think}
Peter Clark, Isaac Cowhey, Oren Etzioni, Tushar Khot, Ashish Sabharwal, Carissa Schoenick, and Oyvind Tafjord. 2018.
\newblock \href {http://arxiv.org/abs/1803.05457} {Think you have solved question answering? try arc, the ai2 reasoning challenge}.

\bibitem[{De{-}Arteaga et~al.(2019)De{-}Arteaga, Romanov, Wallach, Chayes, Borgs, Chouldechova, Geyik, Kenthapadi, and Kalai}]{de-arteaga_bias_2019}
Maria De{-}Arteaga, Alexey Romanov, Hanna~M. Wallach, Jennifer~T. Chayes, Christian Borgs, Alexandra Chouldechova, Sahin~Cem Geyik, Krishnaram Kenthapadi, and Adam~Tauman Kalai. 2019.
\newblock \href {https://doi.org/10.1145/3287560.3287572} {{Bias} in {Bios:} {A} {Case} {Study} of {Semantic} {Representation} {Bias} in a {High-Stakes} {Setting}}.
\newblock In \emph{Proceedings of the Conference on Fairness, Accountability, and Transparency, FAT* 2019, Atlanta, GA, USA, January 29-31, 2019}, pages 120--128. {ACM}.

\bibitem[{Delobelle et~al.(2022)Delobelle, Tokpo, Calders, and Berendt}]{delobelle-etal-2022-measuring}
Pieter Delobelle, Ewoenam Tokpo, Toon Calders, and Bettina Berendt. 2022.
\newblock \href {https://doi.org/10.18653/v1/2022.naacl-main.122} {Measuring fairness with biased rulers: A comparative study on bias metrics for pre-trained language models}.
\newblock In \emph{Proceedings of the 2022 Conference of the North American Chapter of the Association for Computational Linguistics: Human Language Technologies}, pages 1693--1706, Seattle, United States. Association for Computational Linguistics.

\bibitem[{Dubey et~al.(2024)Dubey, Jauhri, Pandey, Kadian, Al-Dahle, Letman, Mathur, Schelten, Yang, Fan, Goyal, Hartshorn, Yang, Mitra, Sravankumar, Korenev, Hinsvark, Rao, Zhang, Rodriguez, Gregerson, Spataru, Roziere, Biron, Tang, Chern, Caucheteux, Nayak, Bi, Marra, McConnell, Keller, Touret, Wu, Wong, Ferrer, Nikolaidis, Allonsius, Song, Pintz, Livshits, Esiobu, Choudhary, Mahajan, Garcia-Olano, Perino, Hupkes, Lakomkin, AlBadawy, Lobanova, Dinan, Smith, Radenovic, Zhang, Synnaeve, Lee, Anderson, Nail, Mialon, Pang, Cucurell, Nguyen, Korevaar, Xu, Touvron, Zarov, Ibarra, Kloumann, Misra, Evtimov, Copet, Lee, Geffert, Vranes, Park, Mahadeokar, Shah, van~der Linde, Billock, Hong, Lee, Fu, Chi, Huang, Liu, Wang, Yu, Bitton, Spisak, Park, Rocca, Johnstun, Saxe, Jia, Alwala, Upasani, Plawiak, Li, Heafield, Stone, El-Arini, Iyer, Malik, Chiu, Bhalla, Rantala-Yeary, van~der Maaten, Chen, Tan, Jenkins, Martin, Madaan, Malo, Blecher, Landzaat, de~Oliveira, Muzzi, Pasupuleti, Singh, Paluri, Kardas, Oldham, Rita,
  Pavlova, Kambadur, Lewis, Si, Singh, Hassan, Goyal, Torabi, Bashlykov, Bogoychev, Chatterji, Duchenne, Çelebi, Alrassy, Zhang, Li, Vasic, Weng, Bhargava, Dubal, Krishnan, Koura, Xu, He, Dong, Srinivasan, Ganapathy, Calderer, Cabral, Stojnic, Raileanu, Girdhar, Patel, Sauvestre, Polidoro, Sumbaly, Taylor, Silva, Hou, Wang, Hosseini, Chennabasappa, Singh, Bell, Kim, Edunov, Nie, Narang, Raparthy, Shen, Wan, Bhosale, Zhang, Vandenhende, Batra, Whitman, Sootla, Collot, Gururangan, Borodinsky, Herman, Fowler, Sheasha, Georgiou, Scialom, Speckbacher, Mihaylov, Xiao, Karn, Goswami, Gupta, Ramanathan, Kerkez, Gonguet, Do, Vogeti, Petrovic, Chu, Xiong, Fu, Meers, Martinet, Wang, Tan, Xie, Jia, Wang, Goldschlag, Gaur, Babaei, Wen, Song, Zhang, Li, Mao, Coudert, Yan, Chen, Papakipos, Singh, Grattafiori, Jain, Kelsey, Shajnfeld, Gangidi, Victoria, Goldstand, Menon, Sharma, Boesenberg, Vaughan, Baevski, Feinstein, Kallet, Sangani, Yunus, Lupu, Alvarado, Caples, Gu, Ho, Poulton, Ryan, Ramchandani, Franco, Saraf,
  Chowdhury, Gabriel, Bharambe, Eisenman, Yazdan, James, Maurer, Leonhardi, Huang, Loyd, Paola, Paranjape, Liu, Wu, Ni, Hancock, Wasti, Spence, Stojkovic, Gamido, Montalvo, Parker, Burton, Mejia, Wang, Kim, Zhou, Hu, Chu, Cai, Tindal, Feichtenhofer, Civin, Beaty, Kreymer, Li, Wyatt, Adkins, Xu, Testuggine, David, Parikh, Liskovich, Foss, Wang, Le, Holland, Dowling, Jamil, Montgomery, Presani, Hahn, Wood, Brinkman, Arcaute, Dunbar, Smothers, Sun, Kreuk, Tian, Ozgenel, Caggioni, Guzmán, Kanayet, Seide, Florez, Schwarz, Badeer, Swee, Halpern, Thattai, Herman, Sizov, Guangyi, Zhang, Lakshminarayanan, Shojanazeri, Zou, Wang, Zha, Habeeb, Rudolph, Suk, Aspegren, Goldman, Damlaj, Molybog, Tufanov, Veliche, Gat, Weissman, Geboski, Kohli, Asher, Gaya, Marcus, Tang, Chan, Zhen, Reizenstein, Teboul, Zhong, Jin, Yang, Cummings, Carvill, Shepard, McPhie, Torres, Ginsburg, Wang, Wu, U, Saxena, Prasad, Khandelwal, Zand, Matosich, Veeraraghavan, Michelena, Li, Huang, Chawla, Lakhotia, Huang, Chen, Garg, A, Silva, Bell,
  Zhang, Guo, Yu, Moshkovich, Wehrstedt, Khabsa, Avalani, Bhatt, Tsimpoukelli, Mankus, Hasson, Lennie, Reso, Groshev, Naumov, Lathi, Keneally, Seltzer, Valko, Restrepo, Patel, Vyatskov, Samvelyan, Clark, Macey, Wang, Hermoso, Metanat, Rastegari, Bansal, Santhanam, Parks, White, Bawa, Singhal, Egebo, Usunier, Laptev, Dong, Zhang, Cheng, Chernoguz, Hart, Salpekar, Kalinli, Kent, Parekh, Saab, Balaji, Rittner, Bontrager, Roux, Dollar, Zvyagina, Ratanchandani, Yuvraj, Liang, Alao, Rodriguez, Ayub, Murthy, Nayani, Mitra, Li, Hogan, Battey, Wang, Maheswari, Howes, Rinott, Bondu, Datta, Chugh, Hunt, Dhillon, Sidorov, Pan, Verma, Yamamoto, Ramaswamy, Lindsay, Lindsay, Feng, Lin, Zha, Shankar, Zhang, Zhang, Wang, Agarwal, Sajuyigbe, Chintala, Max, Chen, Kehoe, Satterfield, Govindaprasad, Gupta, Cho, Virk, Subramanian, Choudhury, Goldman, Remez, Glaser, Best, Kohler, Robinson, Li, Zhang, Matthews, Chou, Shaked, Vontimitta, Ajayi, Montanez, Mohan, Kumar, Mangla, Albiero, Ionescu, Poenaru, Mihailescu, Ivanov, Li, Wang,
  Jiang, Bouaziz, Constable, Tang, Wang, Wu, Wang, Xia, Wu, Gao, Chen, Hu, Jia, Qi, Li, Zhang, Zhang, Adi, Nam, Yu, Wang, Hao, Qian, He, Rait, DeVito, Rosnbrick, Wen, Yang, and Zhao}]{dubey2024llama}
Abhimanyu Dubey, Abhinav Jauhri, Abhinav Pandey, Abhishek Kadian, Ahmad Al-Dahle, Aiesha Letman, Akhil Mathur, Alan Schelten, Amy Yang, Angela Fan, Anirudh Goyal, Anthony Hartshorn, Aobo Yang, Archi Mitra, Archie Sravankumar, Artem Korenev, Arthur Hinsvark, Arun Rao, Aston Zhang, Aurelien Rodriguez, Austen Gregerson, Ava Spataru, Baptiste Roziere, Bethany Biron, Binh Tang, Bobbie Chern, Charlotte Caucheteux, Chaya Nayak, Chloe Bi, Chris Marra, Chris McConnell, Christian Keller, Christophe Touret, Chunyang Wu, Corinne Wong, Cristian~Canton Ferrer, Cyrus Nikolaidis, Damien Allonsius, Daniel Song, Danielle Pintz, Danny Livshits, David Esiobu, Dhruv Choudhary, Dhruv Mahajan, Diego Garcia-Olano, Diego Perino, Dieuwke Hupkes, Egor Lakomkin, Ehab AlBadawy, Elina Lobanova, Emily Dinan, Eric~Michael Smith, Filip Radenovic, Frank Zhang, Gabriel Synnaeve, Gabrielle Lee, Georgia~Lewis Anderson, Graeme Nail, Gregoire Mialon, Guan Pang, Guillem Cucurell, Hailey Nguyen, Hannah Korevaar, Hu~Xu, Hugo Touvron, Iliyan Zarov,
  Imanol~Arrieta Ibarra, Isabel Kloumann, Ishan Misra, Ivan Evtimov, Jade Copet, Jaewon Lee, Jan Geffert, Jana Vranes, Jason Park, Jay Mahadeokar, Jeet Shah, Jelmer van~der Linde, Jennifer Billock, Jenny Hong, Jenya Lee, Jeremy Fu, Jianfeng Chi, Jianyu Huang, Jiawen Liu, Jie Wang, Jiecao Yu, Joanna Bitton, Joe Spisak, Jongsoo Park, Joseph Rocca, Joshua Johnstun, Joshua Saxe, Junteng Jia, Kalyan~Vasuden Alwala, Kartikeya Upasani, Kate Plawiak, Ke~Li, Kenneth Heafield, Kevin Stone, Khalid El-Arini, Krithika Iyer, Kshitiz Malik, Kuenley Chiu, Kunal Bhalla, Lauren Rantala-Yeary, Laurens van~der Maaten, Lawrence Chen, Liang Tan, Liz Jenkins, Louis Martin, Lovish Madaan, Lubo Malo, Lukas Blecher, Lukas Landzaat, Luke de~Oliveira, Madeline Muzzi, Mahesh Pasupuleti, Mannat Singh, Manohar Paluri, Marcin Kardas, Mathew Oldham, Mathieu Rita, Maya Pavlova, Melanie Kambadur, Mike Lewis, Min Si, Mitesh~Kumar Singh, Mona Hassan, Naman Goyal, Narjes Torabi, Nikolay Bashlykov, Nikolay Bogoychev, Niladri Chatterji, Olivier
  Duchenne, Onur Çelebi, Patrick Alrassy, Pengchuan Zhang, Pengwei Li, Petar Vasic, Peter Weng, Prajjwal Bhargava, Pratik Dubal, Praveen Krishnan, Punit~Singh Koura, Puxin Xu, Qing He, Qingxiao Dong, Ragavan Srinivasan, Raj Ganapathy, Ramon Calderer, Ricardo~Silveira Cabral, Robert Stojnic, Roberta Raileanu, Rohit Girdhar, Rohit Patel, Romain Sauvestre, Ronnie Polidoro, Roshan Sumbaly, Ross Taylor, Ruan Silva, Rui Hou, Rui Wang, Saghar Hosseini, Sahana Chennabasappa, Sanjay Singh, Sean Bell, Seohyun~Sonia Kim, Sergey Edunov, Shaoliang Nie, Sharan Narang, Sharath Raparthy, Sheng Shen, Shengye Wan, Shruti Bhosale, Shun Zhang, Simon Vandenhende, Soumya Batra, Spencer Whitman, Sten Sootla, Stephane Collot, Suchin Gururangan, Sydney Borodinsky, Tamar Herman, Tara Fowler, Tarek Sheasha, Thomas Georgiou, Thomas Scialom, Tobias Speckbacher, Todor Mihaylov, Tong Xiao, Ujjwal Karn, Vedanuj Goswami, Vibhor Gupta, Vignesh Ramanathan, Viktor Kerkez, Vincent Gonguet, Virginie Do, Vish Vogeti, Vladan Petrovic, Weiwei Chu,
  Wenhan Xiong, Wenyin Fu, Whitney Meers, Xavier Martinet, Xiaodong Wang, Xiaoqing~Ellen Tan, Xinfeng Xie, Xuchao Jia, Xuewei Wang, Yaelle Goldschlag, Yashesh Gaur, Yasmine Babaei, Yi~Wen, Yiwen Song, Yuchen Zhang, Yue Li, Yuning Mao, Zacharie~Delpierre Coudert, Zheng Yan, Zhengxing Chen, Zoe Papakipos, Aaditya Singh, Aaron Grattafiori, Abha Jain, Adam Kelsey, Adam Shajnfeld, Adithya Gangidi, Adolfo Victoria, Ahuva Goldstand, Ajay Menon, Ajay Sharma, Alex Boesenberg, Alex Vaughan, Alexei Baevski, Allie Feinstein, Amanda Kallet, Amit Sangani, Anam Yunus, Andrei Lupu, Andres Alvarado, Andrew Caples, Andrew Gu, Andrew Ho, Andrew Poulton, Andrew Ryan, Ankit Ramchandani, Annie Franco, Aparajita Saraf, Arkabandhu Chowdhury, Ashley Gabriel, Ashwin Bharambe, Assaf Eisenman, Azadeh Yazdan, Beau James, Ben Maurer, Benjamin Leonhardi, Bernie Huang, Beth Loyd, Beto~De Paola, Bhargavi Paranjape, Bing Liu, Bo~Wu, Boyu Ni, Braden Hancock, Bram Wasti, Brandon Spence, Brani Stojkovic, Brian Gamido, Britt Montalvo, Carl
  Parker, Carly Burton, Catalina Mejia, Changhan Wang, Changkyu Kim, Chao Zhou, Chester Hu, Ching-Hsiang Chu, Chris Cai, Chris Tindal, Christoph Feichtenhofer, Damon Civin, Dana Beaty, Daniel Kreymer, Daniel Li, Danny Wyatt, David Adkins, David Xu, Davide Testuggine, Delia David, Devi Parikh, Diana Liskovich, Didem Foss, Dingkang Wang, Duc Le, Dustin Holland, Edward Dowling, Eissa Jamil, Elaine Montgomery, Eleonora Presani, Emily Hahn, Emily Wood, Erik Brinkman, Esteban Arcaute, Evan Dunbar, Evan Smothers, Fei Sun, Felix Kreuk, Feng Tian, Firat Ozgenel, Francesco Caggioni, Francisco Guzmán, Frank Kanayet, Frank Seide, Gabriela~Medina Florez, Gabriella Schwarz, Gada Badeer, Georgia Swee, Gil Halpern, Govind Thattai, Grant Herman, Grigory Sizov, Guangyi, Zhang, Guna Lakshminarayanan, Hamid Shojanazeri, Han Zou, Hannah Wang, Hanwen Zha, Haroun Habeeb, Harrison Rudolph, Helen Suk, Henry Aspegren, Hunter Goldman, Ibrahim Damlaj, Igor Molybog, Igor Tufanov, Irina-Elena Veliche, Itai Gat, Jake Weissman, James
  Geboski, James Kohli, Japhet Asher, Jean-Baptiste Gaya, Jeff Marcus, Jeff Tang, Jennifer Chan, Jenny Zhen, Jeremy Reizenstein, Jeremy Teboul, Jessica Zhong, Jian Jin, Jingyi Yang, Joe Cummings, Jon Carvill, Jon Shepard, Jonathan McPhie, Jonathan Torres, Josh Ginsburg, Junjie Wang, Kai Wu, Kam~Hou U, Karan Saxena, Karthik Prasad, Kartikay Khandelwal, Katayoun Zand, Kathy Matosich, Kaushik Veeraraghavan, Kelly Michelena, Keqian Li, Kun Huang, Kunal Chawla, Kushal Lakhotia, Kyle Huang, Lailin Chen, Lakshya Garg, Lavender A, Leandro Silva, Lee Bell, Lei Zhang, Liangpeng Guo, Licheng Yu, Liron Moshkovich, Luca Wehrstedt, Madian Khabsa, Manav Avalani, Manish Bhatt, Maria Tsimpoukelli, Martynas Mankus, Matan Hasson, Matthew Lennie, Matthias Reso, Maxim Groshev, Maxim Naumov, Maya Lathi, Meghan Keneally, Michael~L. Seltzer, Michal Valko, Michelle Restrepo, Mihir Patel, Mik Vyatskov, Mikayel Samvelyan, Mike Clark, Mike Macey, Mike Wang, Miquel~Jubert Hermoso, Mo~Metanat, Mohammad Rastegari, Munish Bansal, Nandhini
  Santhanam, Natascha Parks, Natasha White, Navyata Bawa, Nayan Singhal, Nick Egebo, Nicolas Usunier, Nikolay~Pavlovich Laptev, Ning Dong, Ning Zhang, Norman Cheng, Oleg Chernoguz, Olivia Hart, Omkar Salpekar, Ozlem Kalinli, Parkin Kent, Parth Parekh, Paul Saab, Pavan Balaji, Pedro Rittner, Philip Bontrager, Pierre Roux, Piotr Dollar, Polina Zvyagina, Prashant Ratanchandani, Pritish Yuvraj, Qian Liang, Rachad Alao, Rachel Rodriguez, Rafi Ayub, Raghotham Murthy, Raghu Nayani, Rahul Mitra, Raymond Li, Rebekkah Hogan, Robin Battey, Rocky Wang, Rohan Maheswari, Russ Howes, Ruty Rinott, Sai~Jayesh Bondu, Samyak Datta, Sara Chugh, Sara Hunt, Sargun Dhillon, Sasha Sidorov, Satadru Pan, Saurabh Verma, Seiji Yamamoto, Sharadh Ramaswamy, Shaun Lindsay, Shaun Lindsay, Sheng Feng, Shenghao Lin, Shengxin~Cindy Zha, Shiva Shankar, Shuqiang Zhang, Shuqiang Zhang, Sinong Wang, Sneha Agarwal, Soji Sajuyigbe, Soumith Chintala, Stephanie Max, Stephen Chen, Steve Kehoe, Steve Satterfield, Sudarshan Govindaprasad, Sumit Gupta,
  Sungmin Cho, Sunny Virk, Suraj Subramanian, Sy~Choudhury, Sydney Goldman, Tal Remez, Tamar Glaser, Tamara Best, Thilo Kohler, Thomas Robinson, Tianhe Li, Tianjun Zhang, Tim Matthews, Timothy Chou, Tzook Shaked, Varun Vontimitta, Victoria Ajayi, Victoria Montanez, Vijai Mohan, Vinay~Satish Kumar, Vishal Mangla, Vítor Albiero, Vlad Ionescu, Vlad Poenaru, Vlad~Tiberiu Mihailescu, Vladimir Ivanov, Wei Li, Wenchen Wang, Wenwen Jiang, Wes Bouaziz, Will Constable, Xiaocheng Tang, Xiaofang Wang, Xiaojian Wu, Xiaolan Wang, Xide Xia, Xilun Wu, Xinbo Gao, Yanjun Chen, Ye~Hu, Ye~Jia, Ye~Qi, Yenda Li, Yilin Zhang, Ying Zhang, Yossi Adi, Youngjin Nam, Yu, Wang, Yuchen Hao, Yundi Qian, Yuzi He, Zach Rait, Zachary DeVito, Zef Rosnbrick, Zhaoduo Wen, Zhenyu Yang, and Zhiwei Zhao. 2024.
\newblock \href {http://arxiv.org/abs/2407.21783} {The llama 3 herd of models}.

\bibitem[{Eaton(1983)}]{eaton_gauss-markov_1983}
Morris~L. Eaton. 1983.
\newblock \href {https://hdl.handle.net/11299/199431} {The {Gauss}-{Markov} {Theorem} in {Multivariate} {Analysis}}.
\newblock Technical report, University of Minnesota.

\bibitem[{Fu et~al.(2022)Fu, Chen, Lee, and Lee}]{fu_adapterbias_2023}
Chin{-}Lun Fu, Zih{-}Ching Chen, Yun{-}Ru Lee, and Hung{-}yi Lee. 2022.
\newblock \href {https://doi.org/10.18653/v1/2022.findings-naacl.199} {{A}dapter{B}ias: {P}arameter-efficient {T}oken-dependent {R}epresentation {S}hift for {A}dapters in {NLP} {T}asks}.
\newblock In \emph{Findings of the Association for Computational Linguistics: {NAACL} 2022, Seattle, WA, United States, July 10-15, 2022}, pages 2608--2621. Association for Computational Linguistics.

\bibitem[{Gallegos et~al.(2024)Gallegos, Rossi, Barrow, Tanjim, Kim, Dernoncourt, Yu, Zhang, and Ahmed}]{gallegos-etal-2024-bias}
Isabel~O. Gallegos, Ryan~A. Rossi, Joe Barrow, Md~Mehrab Tanjim, Sungchul Kim, Franck Dernoncourt, Tong Yu, Ruiyi Zhang, and Nesreen~K. Ahmed. 2024.
\newblock \href {https://doi.org/10.1162/coli_a_00524} {Bias and fairness in large language models: A survey}.
\newblock \emph{Computational Linguistics}, 50(3):1097--1179.

\bibitem[{Gonen et~al.(2022)Gonen, Ravfogel, and Goldberg}]{gonen-etal-2022-analyzing}
Hila Gonen, Shauli Ravfogel, and Yoav Goldberg. 2022.
\newblock \href {https://doi.org/10.18653/v1/2022.repl4nlp-1.8} {Analyzing gender representation in multilingual models}.
\newblock In \emph{Proceedings of the 7th Workshop on Representation Learning for NLP}, pages 67--77, Dublin, Ireland. Association for Computational Linguistics.

\bibitem[{Hellinger and Bußmann(2002)}]{hellinger2002gender}
Marlis Hellinger and Hadumod Bußmann, editors. 2002.
\newblock \href {https://benjamins.com/catalog/impact.10} {\emph{Gender Across Languages: The Linguistic Representation of Women and Men, Volume 2}}, volume~2 of \emph{Impact: Studies in Language and Society}.
\newblock John Benjamins Publishing Company, Amsterdam/Philadelphia.

\bibitem[{Houlsby et~al.(2019)Houlsby, Giurgiu, Jastrzebski, Morrone, De~Laroussilhe, Gesmundo, Attariyan, and Gelly}]{houlsby2019adapters}
Neil Houlsby, Andrei Giurgiu, Stanislaw Jastrzebski, Bruna Morrone, Quentin De~Laroussilhe, Andrea Gesmundo, Mona Attariyan, and Sylvain Gelly. 2019.
\newblock \href {https://proceedings.mlr.press/v97/houlsby19a.html} {Parameter-efficient transfer learning for {NLP}}.
\newblock In \emph{Proceedings of the 36th International Conference on Machine Learning}, volume~97 of \emph{Proceedings of Machine Learning Research}, pages 2790--2799. PMLR.

\bibitem[{Hu et~al.(2022)Hu, Shen, Wallis, Allen-Zhu, Li, Wang, Wang, and Chen}]{hu2022lora}
Edward~J Hu, Yelong Shen, Phillip Wallis, Zeyuan Allen-Zhu, Yuanzhi Li, Shean Wang, Lu~Wang, and Weizhu Chen. 2022.
\newblock \href {https://openreview.net/forum?id=nZeVKeeFYf9} {Lo{RA}: Low-rank adaptation of large language models}.
\newblock In \emph{International Conference on Learning Representations}.

\bibitem[{Iluz et~al.(2023)Iluz, Limisiewicz, Stanovsky, and Mare{\v{c}}ek}]{iluz-etal-2023-exploring}
Bar Iluz, Tomasz Limisiewicz, Gabriel Stanovsky, and David Mare{\v{c}}ek. 2023.
\newblock \href {https://doi.org/10.18653/v1/2023.ijcnlp-main.57} {Exploring the impact of training data distribution and subword tokenization on gender bias in machine translation}.
\newblock In \emph{Proceedings of the 13th International Joint Conference on Natural Language Processing and the 3rd Conference of the Asia-Pacific Chapter of the Association for Computational Linguistics (Volume 1: Long Papers)}, pages 885--896, Nusa Dua, Bali. Association for Computational Linguistics.

\bibitem[{Kocmi et~al.(2020)Kocmi, Limisiewicz, and Stanovsky}]{kocmi-etal-2020-gender}
Tom Kocmi, Tomasz Limisiewicz, and Gabriel Stanovsky. 2020.
\newblock \href {https://aclanthology.org/2020.wmt-1.39} {Gender coreference and bias evaluation at {WMT} 2020}.
\newblock In \emph{Proceedings of the Fifth Conference on Machine Translation}, pages 357--364, Online. Association for Computational Linguistics.

\bibitem[{Kotek et~al.(2023)Kotek, Dockum, and Sun}]{kotek-etal-2023-gender}
Hadas Kotek, Rikker Dockum, and David~Q. Sun. 2023.
\newblock \href {https://api.semanticscholar.org/CorpusID:261276445} {Gender bias and stereotypes in large language models}.
\newblock \emph{Proceedings of The ACM Collective Intelligence Conference}.

\bibitem[{Levy et~al.(2021)Levy, Lazar, and Stanovsky}]{levy-etal-2021-collecting-large}
Shahar Levy, Koren Lazar, and Gabriel Stanovsky. 2021.
\newblock \href {https://doi.org/10.18653/v1/2021.findings-emnlp.211} {Collecting a large-scale gender bias dataset for coreference resolution and machine translation}.
\newblock In \emph{Findings of the Association for Computational Linguistics: EMNLP 2021}, pages 2470--2480, Punta Cana, Dominican Republic. Association for Computational Linguistics.

\bibitem[{Limisiewicz and Mare{\v{c}}ek(2022)}]{limisiewicz-marecek-2022-dont}
Tomasz Limisiewicz and David Mare{\v{c}}ek. 2022.
\newblock \href {https://doi.org/10.18653/v1/2022.gebnlp-1.3} {Don{'}t forget about pronouns: Removing gender bias in language models without losing factual gender information}.
\newblock In \emph{Proceedings of the 4th Workshop on Gender Bias in Natural Language Processing (GeBNLP)}, pages 17--29, Seattle, Washington. Association for Computational Linguistics.

\bibitem[{Limisiewicz et~al.(2024)Limisiewicz, Mare{\v{c}}ek, and Musil}]{limisiewicz2024debiasing}
Tomasz Limisiewicz, David Mare{\v{c}}ek, and Tom{\'a}{\v{s}} Musil. 2024.
\newblock \href {https://openreview.net/forum?id=XIZEFyVGC9} {Debiasing algorithm through model adaptation}.
\newblock In \emph{The Twelfth International Conference on Learning Representations}.

\bibitem[{Luo et~al.(2023)Luo, Yang, Meng, Li, Zhou, and Zhang}]{Luo2023AnES}
Yun Luo, Zhen Yang, Fandong Meng, Yafu Li, Jie Zhou, and Yue Zhang. 2023.
\newblock \href {https://api.semanticscholar.org/CorpusID:261031244} {An empirical study of catastrophic forgetting in large language models during continual fine-tuning}.
\newblock \emph{ArXiv}, abs/2308.08747.

\bibitem[{Meng et~al.(2022)Meng, Bau, Andonian, and Belinkov}]{meng_locating_2023}
Kevin Meng, David Bau, Alex Andonian, and Yonatan Belinkov. 2022.
\newblock \href {http://papers.nips.cc/paper\_files/paper/2022/hash/6f1d43d5a82a37e89b0665b33bf3a182-Abstract-Conference.html} {{Locating} and {Editing} {Factual} {Associations} in {GPT}}.
\newblock In \emph{NeurIPS}.

\bibitem[{Meng et~al.(2023)Meng, Sharma, Andonian, Belinkov, and Bau}]{meng_mass-editing_2022}
Kevin Meng, Arnab~Sen Sharma, Alex~J. Andonian, Yonatan Belinkov, and David Bau. 2023.
\newblock \href {https://openreview.net/pdf?id=MkbcAHIYgyS} {{Mass-Editing} {Memory} in a {Transformer}}.
\newblock In \emph{The Eleventh International Conference on Learning Representations, {ICLR} 2023, Kigali, Rwanda, May 1-5, 2023}. OpenReview.net.

\bibitem[{Merity et~al.(2016)Merity, Xiong, Bradbury, and Socher}]{merity2016pointer}
Stephen Merity, Caiming Xiong, James Bradbury, and Richard Socher. 2016.
\newblock \href {http://arxiv.org/abs/1609.07843} {Pointer sentinel mixture models}.

\bibitem[{Mitchell et~al.(2022)Mitchell, Lin, Bosselut, Finn, and Manning}]{mitchell2022fast}
Eric Mitchell, Charles Lin, Antoine Bosselut, Chelsea Finn, and Christopher~D Manning. 2022.
\newblock \href {https://openreview.net/pdf?id=0DcZxeWfOPt} {Fast model editing at scale}.
\newblock In \emph{International Conference on Learning Representations}.

\bibitem[{N{\'e}v{\'e}ol et~al.(2022)N{\'e}v{\'e}ol, Dupont, Bezan{\c{c}}on, and Fort}]{neveol-etal-2022-french}
Aur{\'e}lie N{\'e}v{\'e}ol, Yoann Dupont, Julien Bezan{\c{c}}on, and Kar{\"e}n Fort. 2022.
\newblock \href {https://doi.org/10.18653/v1/2022.acl-long.583} {{F}rench {C}row{S}-pairs: Extending a challenge dataset for measuring social bias in masked language models to a language other than {E}nglish}.
\newblock In \emph{Proceedings of the 60th Annual Meeting of the Association for Computational Linguistics (Volume 1: Long Papers)}, pages 8521--8531, Dublin, Ireland. Association for Computational Linguistics.

\bibitem[{OECD(2023)}]{oecd-helth-glance-2023}
OECD. 2023.
\newblock \href {https://doi.org/https://doi.org/https://doi.org/10.1787/7a7afb35-en} {\emph{Health at a Glance 2023}}.

\bibitem[{Patil et~al.(2024)Patil, Hase, and Bansal}]{patil2024can}
Vaidehi Patil, Peter Hase, and Mohit Bansal. 2024.
\newblock \href {https://openreview.net/forum?id=7erlRDoaV8} {Can sensitive information be deleted from {LLM}s? objectives for defending against extraction attacks}.
\newblock In \emph{The Twelfth International Conference on Learning Representations}.

\bibitem[{Popovi{\'c}(2015)}]{popovic-2015-chrf}
Maja Popovi{\'c}. 2015.
\newblock \href {https://doi.org/10.18653/v1/W15-3049} {chr{F}: character n-gram {F}-score for automatic {MT} evaluation}.
\newblock In \emph{Proceedings of the Tenth Workshop on Statistical Machine Translation}, pages 392--395, Lisbon, Portugal. Association for Computational Linguistics.

\bibitem[{Ravfogel et~al.(2020)Ravfogel, Elazar, Gonen, Twiton, and Goldberg}]{ravfogel-etal-2020-null}
Shauli Ravfogel, Yanai Elazar, Hila Gonen, Michael Twiton, and Yoav Goldberg. 2020.
\newblock \href {https://doi.org/10.18653/v1/2020.acl-main.647} {Null it out: Guarding protected attributes by iterative nullspace projection}.
\newblock In \emph{Proceedings of the 58th Annual Meeting of the Association for Computational Linguistics}, pages 7237--7256, Online. Association for Computational Linguistics.

\bibitem[{Ravfogel et~al.(2022)Ravfogel, Twiton, Goldberg, and Cotterell}]{ravfogel_linear_2022}
Shauli Ravfogel, Michael Twiton, Yoav Goldberg, and Ryan Cotterell. 2022.
\newblock \href {https://proceedings.mlr.press/v162/ravfogel22a.html} {{L}inear {A}dversarial {C}oncept {E}rasure}.
\newblock In \emph{International Conference on Machine Learning, {ICML} 2022, 17-23 July 2022, Baltimore, Maryland, {USA}}, volume 162 of \emph{Proceedings of Machine Learning Research}, pages 18400--18421. {PMLR}.

\bibitem[{Rei et~al.(2022)Rei, C.~de Souza, Alves, Zerva, Farinha, Glushkova, Lavie, Coheur, and Martins}]{rei-etal-2022-comet}
Ricardo Rei, Jos{\'e}~G. C.~de Souza, Duarte Alves, Chrysoula Zerva, Ana~C Farinha, Taisiya Glushkova, Alon Lavie, Luisa Coheur, and Andr{\'e} F.~T. Martins. 2022.
\newblock \href {https://aclanthology.org/2022.wmt-1.52} {{COMET}-22: Unbabel-{IST} 2022 submission for the metrics shared task}.
\newblock In \emph{Proceedings of the Seventh Conference on Machine Translation (WMT)}, pages 578--585, Abu Dhabi, United Arab Emirates (Hybrid). Association for Computational Linguistics.

\bibitem[{Saunders and Byrne(2020)}]{saunders-byrne-2020-reducing}
Danielle Saunders and Bill Byrne. 2020.
\newblock \href {https://doi.org/10.18653/v1/2020.acl-main.690} {Reducing gender bias in neural machine translation as a domain adaptation problem}.
\newblock In \emph{Proceedings of the 58th Annual Meeting of the Association for Computational Linguistics}, pages 7724--7736, Online. Association for Computational Linguistics.

\bibitem[{Savoldi et~al.(2021)Savoldi, Gaido, Bentivogli, Negri, and Turchi}]{savoldi_etal_2021}
Beatrice Savoldi, Marco Gaido, Luisa Bentivogli, Matteo Negri, and Marco Turchi. 2021.
\newblock \href {https://doi.org/10.1162/tacl_a_00401} {{Gender Bias in Machine Translation}}.
\newblock \emph{Transactions of the Association for Computational Linguistics}, 9:845--874.

\bibitem[{Stanczak and Augenstein(2021)}]{stanczak_survey_2021}
Karolina Stanczak and Isabelle Augenstein. 2021.
\newblock \href {http://arxiv.org/abs/2112.14168} {{A} {Survey} on {Gender} {Bias} in {Natural} {Language} {Processing}}.
\newblock \emph{CoRR}, abs/2112.14168.

\bibitem[{Stanovsky et~al.(2019)Stanovsky, Smith, and Zettlemoyer}]{stanovsky-etal-2019-evaluating}
Gabriel Stanovsky, Noah~A. Smith, and Luke Zettlemoyer. 2019.
\newblock \href {https://doi.org/10.18653/v1/P19-1164} {Evaluating gender bias in machine translation}.
\newblock In \emph{Proceedings of the 57th Annual Meeting of the Association for Computational Linguistics}, pages 1679--1684, Florence, Italy. Association for Computational Linguistics.

\bibitem[{Touvron et~al.(2023)Touvron, Martin, Stone, Albert, Almahairi, Babaei, Bashlykov, Batra, Bhargava, Bhosale, Bikel, Blecher, Ferrer, Chen, Cucurull, Esiobu, Fernandes, Fu, Fu, Fuller, Gao, Goswami, Goyal, Hartshorn, Hosseini, Hou, Inan, Kardas, Kerkez, Khabsa, Kloumann, Korenev, Koura, Lachaux, Lavril, Lee, Liskovich, Lu, Mao, Martinet, Mihaylov, Mishra, Molybog, Nie, Poulton, Reizenstein, Rungta, Saladi, Schelten, Silva, Smith, Subramanian, Tan, Tang, Taylor, Williams, Kuan, Xu, Yan, Zarov, Zhang, Fan, Kambadur, Narang, Rodriguez, Stojnic, Edunov, and Scialom}]{touvron_llama2_2023}
Hugo Touvron, Louis Martin, Kevin~R. Stone, Peter Albert, Amjad Almahairi, Yasmine Babaei, Nikolay Bashlykov, Soumya Batra, Prajjwal Bhargava, Shruti Bhosale, D.~Bikel, Lukas Blecher, Cristian~Cantón Ferrer, Moya Chen, Guillem Cucurull, David Esiobu, Jude Fernandes, Jeremy Fu, Wenyin Fu, Brian Fuller, Cynthia Gao, Vedanuj Goswami, Naman Goyal, A.~Hartshorn, Saghar Hosseini, Rui Hou, Hakan Inan, Marcin Kardas, Viktor Kerkez, Madian Khabsa, Isabel~M. Kloumann, A.~Korenev, Punit~Singh Koura, Marie-Anne Lachaux, Thibaut Lavril, Jenya Lee, Diana Liskovich, Yinghai Lu, Yuning Mao, Xavier Martinet, Todor Mihaylov, Pushkar Mishra, Igor Molybog, Yixin Nie, Andrew Poulton, Jeremy Reizenstein, Rashi Rungta, Kalyan Saladi, Alan Schelten, Ruan Silva, Eric~Michael Smith, R.~Subramanian, Xia Tan, Binh Tang, Ross Taylor, Adina Williams, Jian~Xiang Kuan, Puxin Xu, Zhengxu Yan, Iliyan Zarov, Yuchen Zhang, Angela Fan, Melanie Kambadur, Sharan Narang, Aurelien Rodriguez, Robert Stojnic, Sergey Edunov, and Thomas Scialom. 2023.
\newblock \href {https://doi.org/10.48550/arXiv.2307.09288} {Llama 2: Open foundation and fine-tuned chat models}.
\newblock \emph{ArXiv}, abs/2307.09288.

\bibitem[{Van Der~Wal et~al.(2022)Van Der~Wal, Jumelet, Schulz, and Zuidema}]{van-der-wal-etal-2022-birth}
Oskar Van Der~Wal, Jaap Jumelet, Katrin Schulz, and Willem Zuidema. 2022.
\newblock \href {https://doi.org/10.18653/v1/2022.gebnlp-1.8} {The birth of bias: A case study on the evolution of gender bias in an {E}nglish language model}.
\newblock In \emph{Proceedings of the 4th Workshop on Gender Bias in Natural Language Processing (GeBNLP)}, pages 75--75, Seattle, Washington. Association for Computational Linguistics.

\bibitem[{Vig et~al.(2020)Vig, Gehrmann, Belinkov, Qian, Nevo, Singer, and Shieber}]{vig_causal_2020}
Jesse Vig, Sebastian Gehrmann, Yonatan Belinkov, Sharon Qian, Daniel Nevo, Yaron Singer, and Stuart~M. Shieber. 2020.
\newblock \href {http://arxiv.org/abs/2004.12265} {{Causal} {Mediation} {Analysis} for {Interpreting} {Neural} {NLP:} {The} {Case} of {Gender} {Bias}}.
\newblock \emph{CoRR}, abs/2004.12265.

\bibitem[{van~der Wal et~al.(2024)van~der Wal, Bachmann, Leidinger, van Maanen, Zuidema, and Schulz}]{vanderwal2023undesirable}
Oskar van~der Wal, Dominik Bachmann, Alina Leidinger, Leendert van Maanen, Willem~H. Zuidema, and Katrin Schulz. 2024.
\newblock \href {https://doi.org/10.1613/JAIR.1.15195} {{U}ndesirable {B}iases in {NLP:} {A}ddressing {C}hallenges of {M}easurement}.
\newblock \emph{J. Artif. Intell. Res.}, 79:1--40.

\bibitem[{Xu et~al.(2024)Xu, Sharaf, Chen, Tan, Shen, Durme, Murray, and Kim}]{xu_contrastive_2024}
Haoran Xu, Amr Sharaf, Yunmo Chen, Weiting Tan, Lingfeng Shen, Benjamin~Van Durme, Kenton Murray, and Young~Jin Kim. 2024.
\newblock \href {http://arxiv.org/abs/2401.08417} {Contrastive preference optimization: Pushing the boundaries of llm performance in machine translation}.

\bibitem[{Zhao et~al.(2018)Zhao, Wang, Yatskar, Ordonez, and Chang}]{zhao-etal-2018-gender}
Jieyu Zhao, Tianlu Wang, Mark Yatskar, Vicente Ordonez, and Kai-Wei Chang. 2018.
\newblock \href {https://doi.org/10.18653/v1/N18-2003} {Gender bias in coreference resolution: Evaluation and debiasing methods}.
\newblock In \emph{Proceedings of the 2018 Conference of the North {A}merican Chapter of the Association for Computational Linguistics: Human Language Technologies, Volume 2 (Short Papers)}, pages 15--20, New Orleans, Louisiana. Association for Computational Linguistics.

\bibitem[{Zmigrod et~al.(2019)Zmigrod, Mielke, Wallach, and Cotterell}]{zmigrod-etal-2019-counterfactual}
Ran Zmigrod, Sabrina~J. Mielke, Hanna Wallach, and Ryan Cotterell. 2019.
\newblock \href {https://doi.org/10.18653/v1/P19-1161} {Counterfactual data augmentation for mitigating gender stereotypes in languages with rich morphology}.
\newblock In \emph{Proceedings of the 57th Annual Meeting of the Association for Computational Linguistics}, pages 1651--1661, Florence, Italy. Association for Computational Linguistics.

\end{thebibliography}
\bibliographystyle{acl_natbib}

\appendix
\newpage
\section{Proofs}
\label{sec:app-theory}

\subsection{Terminological Note}
For brevity of theorems and proofs, we adopt the following notation convention:

\begin{definition}[\textbf{Moore-Penrose Pseudoinvers}]
We denote Moore-Penrose pseudoinverse of matrix $\mat{M}$ as $\mat{M}^\dotplus$:
\begin{equation*}
\mat{M}^\dotplus= (\mat{M}^T\mat{M})^{-1}\mat{M}^T
\end{equation*}
\end{definition}

\begin{definition}[\textbf{Matrix Square Root}]
We denote a positive semi-definite square root of positive semi-definite matrix $\mat{M}$ as $\mat{M}^{1/2}$.

\end{definition}

\begin{definition}[\textbf{Covariance Matrix}]
For two random vectors: $X \in \Real^m$ and $Y \in \Real^n$. We denote the covariance matrix as:
\begin{equation*}
    \mat{\Sigma}_{X,Y} = \cov(X,Y)
\end{equation*}
\end{definition}

\subsection{Proof for DAMA-LEACE Theorem}
We formalize the requirements and implications of that assumption in the following theorem:

\begin{theorem}[\textbf{Gauss-Markov: Probabilistic Least Squares}]
\label{trm:ols}
We consider random vectors: $U$ taking values in $\Real^m$, $V$, and $Z$ taking values in $\Real^n$; both are centered and have finite second moments. We seek the linear regression model given by:
\begin{equation*}
    V = \mat{S}U - \epsilon,
\end{equation*}
given the following assumptions:
\begin{enumerate}
    \item[A] \textbf{No Multicollinearity}: there is no linear relationship among the independent variables, i.e., matrix $\Sigma_{U,U}$ is of full rank $m$.
    \item[B] \textbf{Exogeneity}: the expected value of error terms given independent variables $\E[\epsilon|U]=0$, this also implies that $\cov(\epsilon, U)=0$.
    \item[C] \textbf{Homoscedasticity}: the covariance of the error terms is constant and does not depend on the independent variables $\cov(\epsilon, \epsilon|U) = \sigma\Iden$.
\end{enumerate}

Then, the ordinary least squares estimator is given by the formula:
\begin{equation*}
    \mat{S}^* = \mat{\Sigma}_{U,V}\mat{\Sigma}_{U,U}^{-1}
\end{equation*}
Such estimator is \textbf{best linear unbiased estimator} and minimizes the variance of error terms: $Tr(\cov(\epsilon, \epsilon))$.
\end{theorem}

The proof of the Theorem~\ref{trm:ols} can be found in the classical statistics literature.
For instance, \citet{eaton_gauss-markov_1983} presents proof for the multivariate case presented above.

Equipped with the theorems above, we are ready to present the theorem that is the main theoretical contribution of this work:

\begin{reptheorem}{trm:dama-leace}
We consider random vectors: $U$ taking values in $\Real^m$, $V$ and $Z$ taking values in $\Real^n$, where $m \geq n$. Under assumptions that: A) random vectors $U$, $V$, $Z$ are centered, and each of them has finite moment; B) the regression relation between $U$ and $V$ fulfill the assumption of ordinary least squares, and there exist least squares estimator $V = \mat{S}U - \epsilon$.

Then the objective:
\begin{equation*}
    \argmin_{\mat{P} \in \Real{n \times m}} \E\left[|| \mat{P}U -V||^2\right],
\end{equation*}
subject to:
\begin{equation*}
    \cov(\mat{P}U,Z)=0
\end{equation*}
is solved by:
\begin{equation*}
    \mat{P}^* = \left(\Iden - \mat{W}^\dotplus \mat{P}_{\mat{W}\mat{\Sigma}}\mat{W}\right)\mat{S},
\end{equation*}
where $\mat{W}$ is the whitening transformation $(\Sigma^{1/2}_{\mat{S}U,\mat{S}U})^\dotplus$;  $\mat{P}_{\mat{W}\mat{\Sigma}}$ is an orthogonal projection matrix onto colspace of $\mat{W}\mat{\Sigma}_{\mat{S}U,Z}$; $\mat{S}$ is a least squares estimator of $V$ given $U$:  $\mat{S}=\mat{\Sigma}_{U,V}\mat{\Sigma}_{U,U}^{-1}$.
\end{reptheorem}


\begin{proof}

For simplicity, we will decompose the problem into independent optimization objectives corresponding to each dimension in $\Real^n$. For the $i$th dimension, we write the objective as:

\begin{equation}
\label{eqn:ith-objective}
\argmin_{\mat{P}_i \in \Real^n}\E\left[\mat{P}_i^T V - V_i\right]^2 \quad \text{s.t.} \quad \cov(\mat{P}_i U, Z) = 0,
\end{equation}
where $\mat{P}_i$ is $i$th column of matrix $\mat{P}$.
From the assumption (B) of the theorem, we can represent the linear relation between $U$ and $V$, as $\mat{S}U = V + \epsilon$, where $\epsilon$ is an error term of regression. We use this property to rewrite the minimization objective from expression~\ref{eqn:ith-objective}, as:

\begin{equation}
\label{eqn:double-objective}
\argmin_{\widetilde{\mat{P}_i} \in \Real^n, \mat{S} \in \Real^{m \times n}} \E\left[\widetilde{\mat{P}_i}^T\mat{S} U - V_i\right]^2
\end{equation}

We manipulate the term under $\argmin$ to rewrite it as a sum of three terms:

\begin{equation}
\label{eqn:objectives-decomposed}
\begin{split}
\E\left[\widetilde{\mat{P}_i}^T\mat{S} U - V_i\right]^2 
= \E\left[\widetilde{\mat{P}_i}^T(V + \epsilon) - V_i\right]^2 = \\
= \E\left[\widetilde{\mat{P}_i}^T(V + \epsilon) - (V_i + \epsilon_i) + \epsilon_i \right]^2 = \\
= 
\underbrace{2 E\left[ \left(\widetilde{\mat{P}_i}^T(V + \epsilon)  - (V_i + \epsilon_i)\right)\epsilon_i\right]}_{\text{I}} + \\
+ \underbrace{\E[\epsilon_i]^2}_{\text{II}} 
+ \underbrace{\E\left[\widetilde{\mat{P}_i}^T(V + \epsilon)  - (V_i + \epsilon_i) \right]^2}_{\text{III}}
\end{split}
\end{equation}

We will now consider each of the three summands one by one to find the solution to the optimization objective $P^* = \widetilde{P}^* S^*$.


\paragraph{Summand I} zeros out. We show that by observing that the summand is doubled covariance\footnote{From the fact that both factors under $\E$ are centered.}:

\begin{equation}
\begin{split}
E\left[ \left(\widetilde{\mat{P}_i}^T(V + \epsilon)  - (V_i + \epsilon_i)\right)\epsilon_i\right] = \\
= \cov\left(\widetilde{\mat{P}_i}^T(V + \epsilon)  - (V_i + \epsilon_i), \epsilon_i\right) = \\
= \left(\widetilde{\mat{P}_i}^T - \mathds{1}_i^T\right)\cov(V-\epsilon, \epsilon) = \\
= \left(\widetilde{\mat{P}_i}^T - \mathds{1}_i^T\right) \mat{S} \cov(U, \epsilon)
\end{split}
\end{equation}

From assumption B of Theorem~\ref{trm:ols} (exogeneity) and, by extension, assumption of this theorem, we have that $\cov(U, \epsilon) = 0$ and thus summand I zeros out.

\paragraph{Summand II} by the conclusion of Theorem~\ref{trm:ols} is minimized by setting:
\begin{equation}
\mat{S}^* = \mat{\Sigma_{U,V}}\mat{\Sigma_{U,U}}^{-1}
\end{equation}
We can also set $\mat{S}$ to $\mat{S}^*$ in summand III, as the variable under $\E$ is independent of $\epsilon$, as shown in the previous paragraph. By finding $\mat{S}^*$, we have solved part of the objective in expression~\ref{eqn:double-objective}.

\paragraph{Summand III} we find the matrix $\widetilde{P}$ minimizing the value of the summand under constraines. By rewriting $\cov(P_i U), Z)$ as $\cov(\widetilde{\mat{P}_i}(V + \epsilon, Z)$, we observe that minimizing the value of summand III under constraint is analogical to solving the problem stated in LEACE (Theorem~\ref{trm:leace}):
\begin{equation}
\begin{split}
\label{eqn:leace-analogy}
\argmin_{\widetilde{\mat{P}_i} \in \Real^n}\E\left[\widetilde{\mat{P}_i}^T (V + \epsilon) - (V_i + \epsilon)\right]^2 \\
\text{such that} \quad \cov(\widetilde{\mat{P}}_i (V + \epsilon), Z) = 0
\end{split}
\end{equation}

We find the solution based on Theorem~\ref{trm:leace}, where we substitute $X$ with $V + \epsilon$ and find $\widetilde{\mat{P}}^* = \Iden - \mat{W}^\dotplus \mat{P}_{\mat{W}\mat{\Sigma}}\mat{W}$, where $\mat{W}$ is the whitening transformation $(\Sigma^{1/2}_{V+\epsilon,V+\epsilon})^\dotplus$;  $\mat{P}_{\mat{W}\mat{\Sigma}}$ is an orthogonal projection matrix onto colspace of $\mat{W}\mat{\Sigma}_{V+\epsilon,Z}$

\paragraph{Conclusion} for summands II and III, we independently found the matrices minimizing their values. We obtain the matrix $P^*$ solving our original objective in expression~\ref{eqn:ith-objective} by multiplying them:
\begin{equation}
P^* = \widetilde{P}^* S^* = \left(\Iden - \mat{W}^\dotplus \mat{P}_{\mat{W}\mat{\Sigma}}\mat{W}\right)\mat{\Sigma}_{U,V}\mat{\Sigma}_{U,U}^{-1}
\end{equation}
\end{proof}

\subsection{Proof for Dual-Debiasing Theorem}
\begin{reptheorem}{trm:dual-debiasing}
We consider random vectors $X$, $Z_b$, and $Z_f$ in $\Real^n$. 
Under the assumptions that:
A)  $Z_b$ and $Z_f$ $Z_b \perp Z_f | X$, i.e., $Z_b$ and $Z_f$ are conditionally independent, given $X$;
B) $\mat{\Sigma}_{X,Z_b}\mat{\Sigma}_{X,Z_f}^T$, i.e., the variable $X$ is correlated with $Z_f$ and $Z_b$ through mutually orthogonal subspaces of $\Real^n$.
The solution of the objective:
\begin{equation*}
    \argmin_{\mat{P} \in \Real^{n \times n}} \E\left[|| \mat{P}X -X||^2\right],
\end{equation*}
subject to:
\begin{equation*}
    \cov(\mat{P}X,Z_b)=0,
\end{equation*}
satisfies:
\begin{equation*}
    \cov(\mat{P}X, Z_f)=\cov(X,Z_f).
\end{equation*}
\end{reptheorem}

\begin{proof}

First, we observe that the assumption A) can be generalized to any coordinate system. For an orthogonal matrix $\mat{W}$, we have:
\begin{equation}
\label{eqn:w-generalization}
\mat{\Sigma}_{\mat{W}X,Z_b}\mat{\Sigma}_{\mat{W}X,Z_f}^T 
=\mat{W}\mat{\Sigma}_{X,Z_b}\mat{\Sigma}_{X,Z_f}^T \mat{W}^T = 0
\end{equation}
This guarantees the orthogonality of spaces spanned by columns of two orthogonality matrices.
The property will be useful for the second part of the proof:
\begin{equation}
\label{eqn:colspace-orthogonality}
Col(\mat{\Sigma}_{\mat{W}X,Z_b}) \perp Col(\mat{\Sigma}_{\mat{W}X,Z_f})
\end{equation}

Secondly, we remind the reader that the solution to the objective provided in the theorem (based on Theorem~\ref{trm:leace}) is as follows:
\begin{equation}
\mat{P}^* = \Iden - \mat{W}^\dotplus \mat{P}_{\mat{W}\mat{\Sigma}}\mat{W}
\end{equation}
Now, we evaluate the covariance matrix between $\mat{P}^* X$ and $Z_f$ to check that it is the same as the covariance matrix between $X$ and $Z_f$.
\begin{equation}
\begin{split}
\label{eqn:dual-leace-application}
\cov(\mat{P}^* X, Z_f) = \mat{\Sigma}_{X,Z_f} - \mat{W}^\dotplus \mat{P}_{\mat{W}\mat{\Sigma}}\mat{W} \mat{\Sigma}_{X,Z_f} = \\
= \mat{\Sigma}_{X,Z_f} - \mat{W}^\dotplus \mat{P}_{\mat{W}\mat{\Sigma}} \mat{\Sigma}_{\mat{W}X,Z_f}
\end{split}
\end{equation}
we note that $\mat{P}_{\mat{W}\mat{\Sigma}}$ is the projection matrix onto the column space of $\mat{\Sigma}_{\mat{W}X,Z_f}$.
From that fact and Equation~\ref{eqn:colspace-orthogonality}, we have:
\begin{equation}
\mat{P}_{\mat{W}\mat{\Sigma}} \mat{\Sigma}_{\mat{W}X,Z_f} = 0
\end{equation}
Thus the last component in Equation~\ref{eqn:dual-leace-application} nullifies and 
we conclude that:
\begin{equation}
\cov(\mat{P}^* X, Z_f) = \mat{\Sigma}_{X,Z_f} = \cov(X, Z_f) 
\end{equation}
\end{proof}

\section{Prompts}
\label{sec:app-prompts}

\subsection{Monolingual Prompts}
\label{sec:monolingual-prompts}
The list of 11 prompt templates is given in Table~\ref{tab:prompt-templates-en}. 
The term <profession> is substituted by 219 professions  without factual gender (from \citealp{bolukbasi_2016}) and 26 gendered entities,
\footnote{``\textit{man}'', ``\textit{boy}'', ``\textit{gentleman}'', ``\textit{father}'', ``\textit{son}'', ``\textit{brother}'', ``\textit{husband}'', ``\textit{king}'', ``\textit{prince}'', ``\textit{uncle}'', ``\textit{nephew}'', ``\textit{groom}'', ``\textit{duke}'', ``\textit{grandfather}'', ``\textit{widower}'', ``\textit{hero}'', ``\textit{wizard}'', ``\textit{baron}'', ``\textit{emperor}'', ``\textit{anchorman}'', ``\textit{clergyman}'', ``\textit{chairman}'', ``\textit{alumnus}'', ``\textit{woman}'', ``\textit{girl}'', ``\textit{lady}'', ``\textit{mother}'', ``\textit{daughter}'', ``\textit{sister}'', ``\textit{wife}'', ``\textit{queen}'', ``\textit{princess}'', ``\textit{aunt}'', ``\textit{niece}'', ``\textit{bride}'', ``\textit{duchess}'', ``\textit{grandmother}'', ``\textit{widow}'', ``\textit{heroine}'', ``\textit{witch}'', ``\textit{baroness}'', ``\textit{empress}'', ``\textit{anchorwoman}'', ``\textit{clergywomen}'', ``\textit{chairwomen}'', ``\textit{alumna}''}
which makes 2409 stereotypical and 286 factual prompts in total used for \dd{}.

\subsection{Multilingual Prompts}
\label{sec:multilingual-prompts}
For debiasing the translation models, we use 11 English-German and 11 English-Czech prompt templates shown in Table~\ref{tab:prompt-templates-ende-encs}. 
These prompts were designed to be as general as possible, and so that the target language prefix does not include any gender information, while the proposed completions, which are always one-token words, do specify the gender. In German, these completions may be determiners. Czech does not use determiners, but gender is included, for example, in the past form of the verb ``to be''.
We use the same sets of professions, and therefore, we generate additional data of equal size for each language.
To match the intended use, we wrap them into ALMA-R translation prompt template:

\vspace{2mm}
\noindent Translate this from \{src\_lang\} to \{tgt\_lang\}:\\
\{src\_lang\}: <English source>\\
\{tgt\_lang\}: <Target prefix>\\

\begin{table}[ht]
\begin{center}\
\footnotesize
\begin{tabular}{ll}
\toprule
English prompt & Completions \\
\midrule
The <profession> wanted that & [he, she, they] \\
The <profession> laughed because & [he, she, they] \\
The <profession> went home because & [he, she, they] \\
The <profession> desired that & [he, she, they] \\
The <profession> wished that & [he, she, they] \\
The <profession> cried because & [he, she, they] \\
The <profession> ate because & [he, she, they] \\
The <profession> said that & [he, she, they] \\
The <profession> ran because & [he, she, they] \\
The <profession> stayed up because & [he, she, they] \\
The <profession> whispered that & [he, she, they] \\
\bottomrule
\end{tabular}
\end{center}
\caption{Monolingual English prompt templates.}
\label{tab:prompt-templates-en}
\end{table}

\begin{table*}[t]
\begin{center}\
\footnotesize
\begin{tabular}{lll}
\toprule
English source & German prefix & Completions \\
\midrule
This is the <profession>. & Das ist \_\_\_ & [der, die] \\
There is the <profession>. & Da ist \_\_\_ & [der, die] \\
The <profession> is not working today. & \_\_\_ & [Der, Die] \\
The <profession> was fired. & \_\_\_ & [Der, Die] \\
The <profession> is busy. & \_\_\_ & [Der, Die] \\
Do you know the <profession> & Kennen Sie \_\_\_ & [den, die] \\
I was there with the <profession> & Ich war dort mit \_\_\_ & [dem, der] \\
I asked the <profession>. & Ich fragte \_\_\_ & [den, die] \\
We met the <profession>. & Wir trafen \_\_\_ & [den, die] \\
I answered the <profession>. & Ich antwortete \_\_\_ & [dem, der] \\
The salary of the <profession> has increased. &  Das Gehalt \_\_\_ & [des, der] \\
\bottomrule
\end{tabular}
\begin{tabular}{lll}
\toprule
English source & Czech prefix & Completions \\
\midrule
This is that <profession>. & To je \_\_\_ & [ten, ta] \\
There is that <profession>. & Tam je \_\_\_ & [ten, ta] \\
That <profession> is not working today. & \_\_\_ & [Ten, Ta] \\
That <profession> was fired. & \_\_\_ & [Ten, Ta] \\
That <profession> is busy. & \_\_\_ & [Ten, Ta] \\
I was a <profession> two years ago. & Před dvěma lety jsem \_\_\_ & [byl, byla] \\
You were a <profession> two years ago. & Před dvěma lety jste \_\_\_ & [byl, byla] \\
If only I were a <profession>. & Kdybych tak \_\_\_ & [byl, byla] \\
I was a <profession> at that time. & V té době jsem \_\_\_ & [byl, byla] \\
You were a <profession> at that time. & V té době jsi \_\_\_ & [byl, byla] \\
You were a <profession> at that time. & V té době jste \_\_\_ & [byl, byla] \\
\bottomrule
\end{tabular}
\end{center}
\caption{Multilingual prompt templates for English-to-German and English-to-Czech translation}
\label{tab:prompt-templates-ende-encs}
\end{table*}



\section{Additional Results}
\label{sec:app-results}

\begin{figure*}[tb]
     \centering
     \begin{subfigure}[b]{0.4\textwidth}
         \centering
         \includegraphics[width=\textwidth]{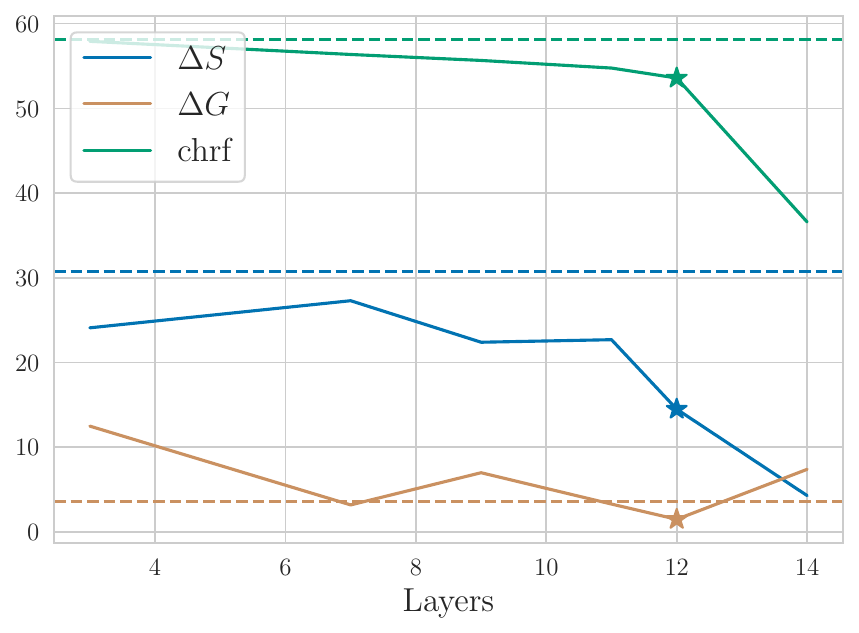}
         \caption{Bias-to-feature threshold fixed at 0.05}
     \end{subfigure}
     \begin{subfigure}[b]{0.4\textwidth}
         \centering
         \includegraphics[width=\textwidth]{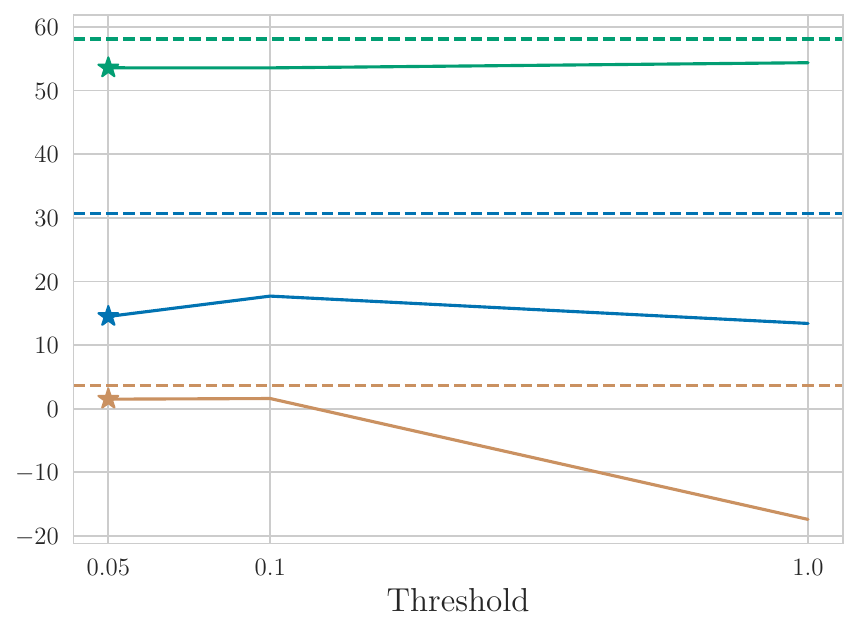}
         \caption{Number of layers fixed at 12}
     \end{subfigure}
     
        \caption{The hyperparameter analysis for \ddama{} applied to ALMA-R 13B model on performance and bias in translation to German. We measured bias via WinoMT metrics $\Delta S$ and $\Delta G$. The translation quality to Germna is measured by chrf on WMT-22. Stars mark the performance of the best setting. The dashed line corresponds to the scores of the original model.}
        \label{fig:abaltion_alamr}
\end{figure*}

\subsection{Stereotypical and Factual Signals across Layers}

\begin{figure}[!tb]
     \centering
     \begin{subfigure}[b]{\linewidth}
        \includegraphics[width=\linewidth]{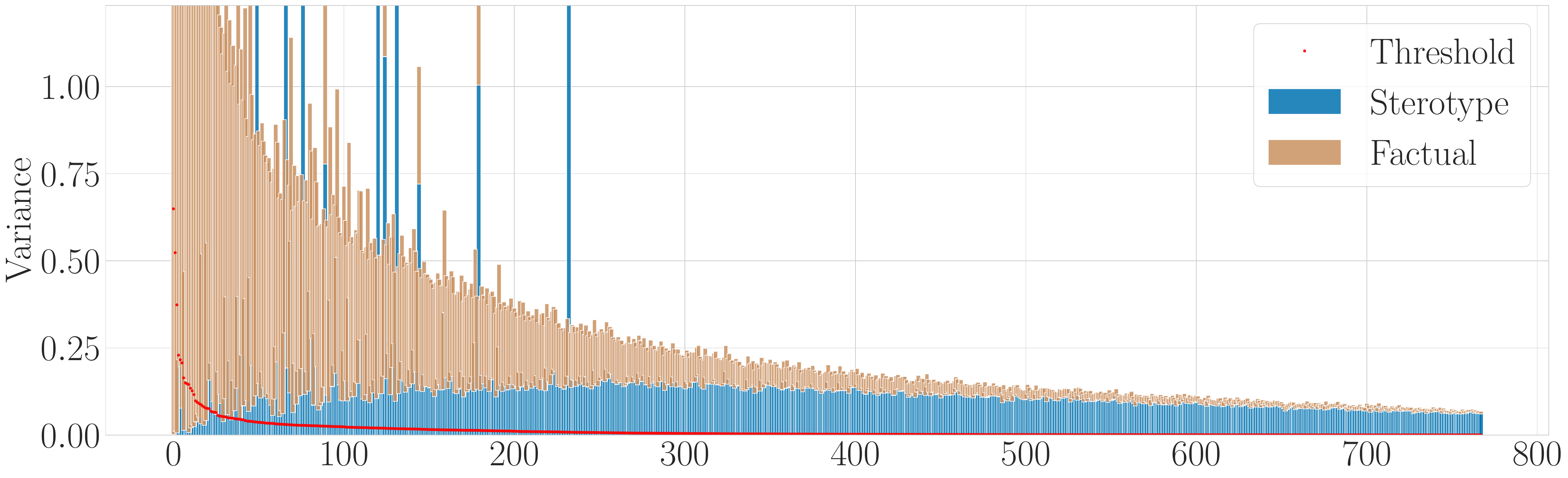}
         \caption{Layer 28}
     \end{subfigure}
     \begin{subfigure}[b]{\linewidth}
         \centering
         \includegraphics[width=\linewidth]{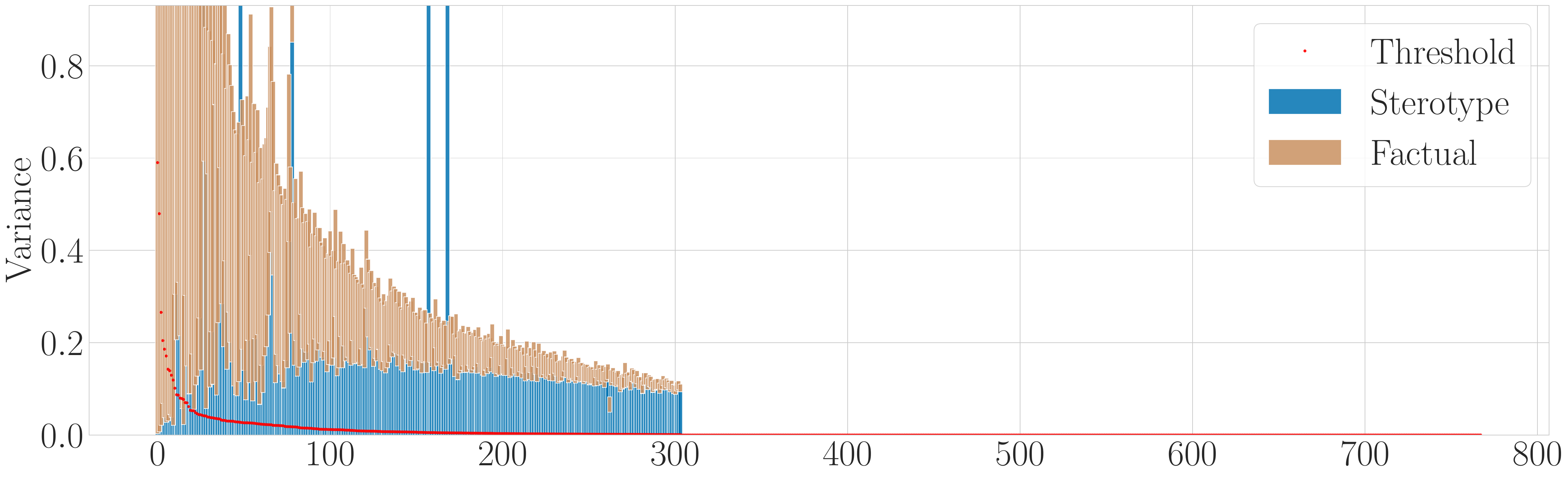}
         \caption{Layer 31}
     \end{subfigure}
     \begin{subfigure}[b]{\linewidth}
         \centering
         \includegraphics[width=\linewidth]{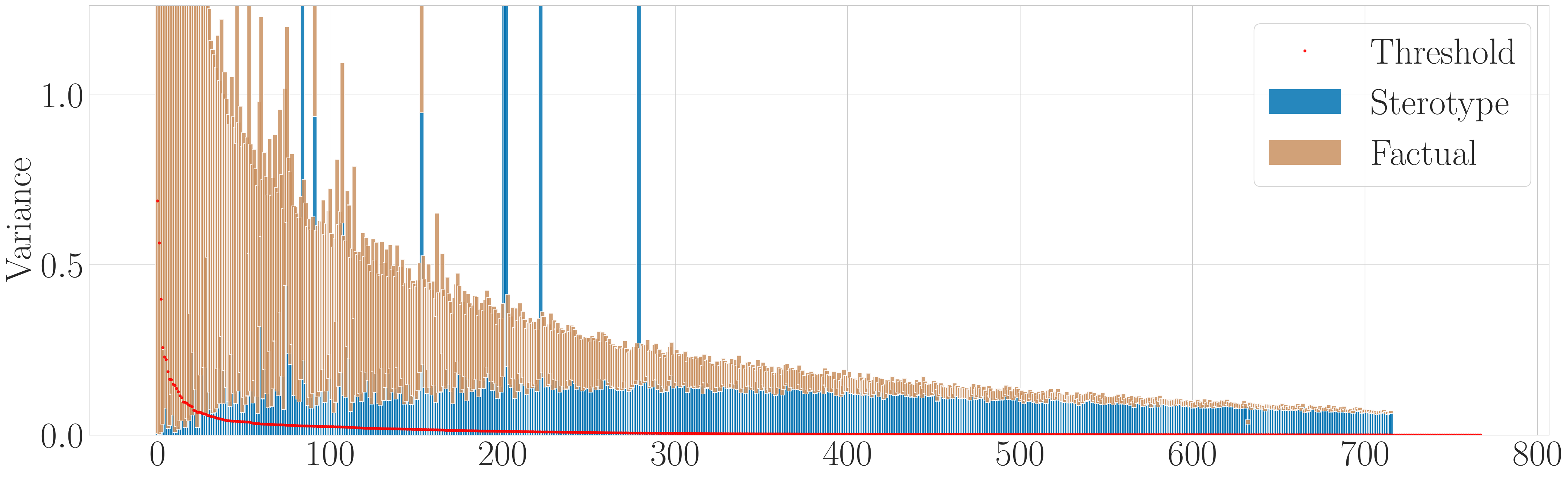}
         \caption{Layer 34}
     \end{subfigure}
     \begin{subfigure}[b]{\linewidth}
         \centering
         \includegraphics[width=\linewidth]{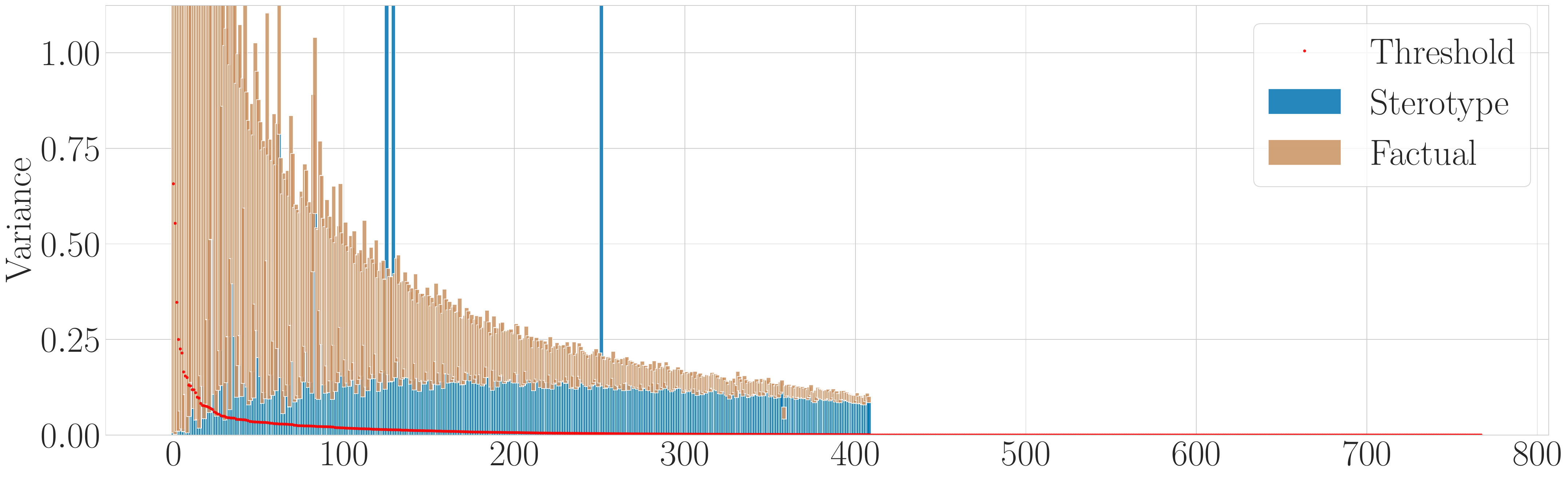}
         \caption{Layer 37}
     \end{subfigure}
    \caption{Visualization of dimensions and their variances related to stereotypical and factual gender signals identified by \dd{} algorithm across different layers of \llama{}~2 13B.}
    \label{fig:concept-erasure-across-layers}
\end{figure}

In Figure~\ref{fig:concept-erasure-across-layers}, we observe the variances with stereotypical and factual gender signals in subsequent layers.
We see that the number of biased dimensions differs across layers.
Nevertheless, we observe the same pattern in each layer: stereotypical signal is encoded in a relatively small number of dimensions with high variance, while the stereotypical variance is spread across more dimensions with lower values in each.

\subsection{Choice of Hyperparameters in Translation}
\label{sec:hyperparameters-translation}

Analogically to Section~\ref{sec:hyperparameters}, we present the impact of bias-to-feature threshold $t$ and the number of edited layers on translation to German in Figure~\ref{fig:abaltion_alamr}.
We observe that stronger factual regularization (high $t$) helps in reducing 
representational bias ($\Delta G$) yet offers weaker stereotypical bias mitigation ($\Delta S$).
Similar to the results in language modeling, the best performance is obtained when editing 12 mid-upper layers with $t=0.05$.

\end{document}